\documentclass{article}

\usepackage[nonatbib,preprint]{neurips_2020}

\usepackage[utf8]{inputenc} %
\usepackage[T1]{fontenc}    %
\usepackage{hyperref}       %
\usepackage{url}            %
\usepackage{booktabs}       %
\usepackage{amsfonts}       %
\usepackage{nicefrac}       %
\usepackage{microtype}      %
\usepackage{graphicx}
\usepackage{xcolor}
\usepackage{amsmath}
\usepackage{cleveref}

\newcommand{\X}{\mathcal{X}}
\newcommand{\Y}{\mathcal{Y}}
\renewcommand{\L}{\mathcal{L}}
\newcommand{\y}{\mathbf{y}}

\newcommand{\dy}{\delta \y}
\renewcommand{\v}{\mathbf{v}}
\renewcommand{\u}{\mathbf{u}}

\newcommand{\E}{\mathbb{E}}

\DeclareMathOperator{\cov}{cov}

\usepackage{lipsum}
\usepackage{soul}
\usepackage{enumerate}

\newcommand{\lin}{\text{lin}}

\newcommand{\NTK}{\Theta}

\newtheorem{prop}{Proposition}

\usepackage{subcaption}
\usepackage{varwidth}
\usepackage[ruled,vlined]{algorithm2e}
\usepackage{algorithmic}

\title{%
Predicting Training Time Without Training %
}

\author{%
  Luca~Zancato$^{1,2}$ \  Alessandro~Achille$^2$ \  Avinash~Ravichandran$^2$ \  Rahul~Bhotika$^2$ \  Stefano~Soatto$^2$\\
  University of Padova$^1$ \quad \quad Amazon Web Services$^2$\\
  \texttt{luca.zancato@phd.unipd.it} \quad \texttt{\{aachille,ravinash,bhotikar,soattos\}@amazon.com}\\
}

\begin{document}

\maketitle

\begin{abstract}
We tackle the problem of predicting the number of optimization steps that a pre-trained deep network needs to converge to a given value of the loss function. To do so, we leverage the fact that the training dynamics of a deep network during fine-tuning are well approximated by those of a  linearized model. This allows us to approximate the training loss and accuracy at any point during training by solving a low-dimensional Stochastic Differential Equation (SDE) in function space. Using this result, we are able to predict the time it takes for Stochastic Gradient Descent (SGD) to fine-tune a model to a given loss without having to perform any training.
In our experiments, we are able to predict training time of a ResNet within a 20\% error margin on a variety of datasets and hyper-parameters, at a 30 to 45-fold reduction in cost compared to actual training. We also discuss how to further reduce the computational and memory cost of our method, and in particular we show that by exploiting the spectral properties of the gradients' matrix it is possible predict training time on a large dataset while processing only a subset of the samples.
\end{abstract}

\section{Introduction}
Say you are a researcher with many more ideas than available time and compute resources to test them.
You are pondering to launch thousands of experiments
but, as the deadline approaches, you wonder whether they will finish in time, and before your computational budget is exhausted.
Could you predict the time it takes for a network to converge, before even starting to train it?

We look to efficiently estimate the number of training steps a Deep Neural Network (DNN) needs to converge to a given value of the loss function, without actually having to train the network. This problem has received little attention thus far, possibly due to the fact that the initial training dynamics of a randomly initialized DNN are highly non-trivial to characterize and analyze. However, in most practical applications, it is common to {\em not} start from scratch, but from a pre-trained model. This may simplify the analysis, since the final solution obtained by fine-tuning is typically not too far from the initial solution obtained after pre-training. In fact, it is known that the dynamics of overparametrized DNNs \cite{pmlr-v97-du19c, DBLP:journals/corr/abs-1811-08888, allen2018convergence} during fine-tuning tends to be more predictable and close to convex \cite{mu2020gradients}.

We therefore characterize the training dynamics of a pre-trained network and provide a computationally efficient procedure to estimate the expected profile of the loss curve over time. In particular, we provide qualitative interpretation and quantitative prediction of the convergence speed of a DNN as a function of the network pre-training, the target task, and the optimization hyper-parameters.

We use a linearized version of the DNN model around pre-trained weights to study its actual dynamics. In \cite{lee2019wide} a similar technique is used to describe the learning trajectories of \textit{randomly initialized} wide neural networks. Such an approach is inspired by the Neural Tangent Kernel (NTK)  for infinitely wide networks \cite{jacot2018neural}. While we note that NTK theory may not correctly predict the dynamics of real (finite size) randomly initialized networks \cite{goldblum2019truth}, we show that our linearized approach can be extended to fine-tuning of real networks in a similar vein to \cite{mu2020gradients}. In order to predict fine-tuning Training Time (TT) without training we introduce a Stochastic Differential Equation (SDE) (similar to \cite{hayou2019mean}) to approximate the behavior of SGD: we do so for a linearized DNN and in function space rather than in weight space. That is, rather than trying to predict the evolution of the weights of the network (a $D$-dimensional vector), we aim to predict the evolution of the outputs of the network on the training set (a $N\times C$-dimensional vector, where $N$ is the size of the dataset and $C$ the number of network's outputs).
This drastically reduces the dimensionality of the problem for over-parametrized networks (that is, when $NC \ll D$).

A possible limiting factor of our approach is that the memory requirement to predict the dynamics scales as $O(D C^2 N^2)$. This would rapidly become infeasible for datasets of moderate size and for real architectures ($D$ is in the order of millions). To mitigate this, we show that we can use random projections to restrict to a much smaller $D_0$-dimensional subspace with only minimal loss in prediction accuracy.
We also show how to estimate Training Time using a small subset of $N_0$ samples, which reduces the total complexity to $O(D_0\,C^2N_0^2)$. We do this by exploiting the spectral properties of the Gram matrix of the gradients. Under mild assumptions the same tools can be used to estimate Training Time on a larger dataset without actually seeing the data.

To summarize, our main contributions are:
\begin{enumerate}[(i)]
    \item We present both a qualitative and quantitative analysis of the fine-tuning Training Time as a function of the Gram-Matrix $\Theta$ of the gradients at initialization (empirical NTK matrix).
    \item We show how to reduce the cost of estimating the matrix $\Theta$ using random projections of the gradients, which makes the method efficient for common architectures and large datasets.
    \item We introduce a method to estimate how much longer a network will need to train if we increase the size of the dataset without actually having to see the data (under the hypothesis that new data is sampled from the same distribution).
    \item We test the accuracy of our predictions on off-the-shelf state-of-the-art models trained on real
    datasets. We are able to predict the correct training time within a 20\% error with 95\% confidence over several different datasets and hyperparameters at only a small fraction of the time it would require to actually run the training (30-45x faster in our experiments).
\end{enumerate}

\section{Related Work}
Predicting the training time of a state-of-the-art architecture on large scale datasets is a relatively understudied topic.
In this direction, Justus et al. \cite{DBLP:journals/corr/abs-1811-11880} try to estimate the wall-clock time required for a forward and backward pass on given hardware. We focus instead on a complementary aspect:  estimating the number of fine-tuning steps necessary for the loss to converge below a given threshold. Once this  has been estimated we can combine it with the average time for the forward and backward pass to get a final estimate of the wall clock time to fine-tune a DNN model without training it.

Hence, we are interested in predicting the learning dynamics of a pre-trained DNN trained with either Gradient Descent (GD) or Stochastic Gradient Descent (SGD).
While different results are known to describe training dynamics under a variety of assumptions (e.g. \cite{DBLP:journals/corr/KeskarMNST16, Smith2018ABP, saxe2013exact, DBLP:journals/corr/abs-1710-10174}), in the following we are mainly interested on recent developments which describe the optimization dynamics of a DNN using a linearization approach. Several works \cite{jacot2018neural, lee2017deep,du2018gradient_theory_1} suggest that in the over-parametrized regime wide DNNs behave similar to linear models, and in particular they are fully characterized by the Gram-Matrix of the gradients, also known as empirical Neural Tangent Kernel (NTK).

Under these assumptions, \cite{jacot2018neural,arora2019fine} derive a simple  connection between training time and spectral decomposition of the NTK matrix. However, their results are limited to Gradient Descend dynamics and to simple architectures which are not directly applicable to real scenarios. In particular, their arguments hinge on the assumption of using a randomly initialized very wide two-layer or infinitely wide neural network \cite{arora2019fine, du2018gradient, DBLP:journals/corr/LiY17c}. We take this direction a step further, providing a unified framework which allows us to describe training time for both SGD and  GD on common architectures.

Again, we rely on a linear approximation of the model, but while the practical validity of such linear approximation for randomly initialized state-of-the-art architectures (such as ResNets) is still discussed \cite{goldblum2019truth}, we follow Mu et al. \cite{mu2020gradients} and argue that the fine-tuning dynamics of over-parametrized DNNs can be closely described by a linearization. We expect such an approximation to hold true since the network
 does not move much in parameters space during fine-tuning and \textit{over-parametrization} leads to smooth and regular loss function around the pre-trained weights \cite{pmlr-v97-du19c, DBLP:journals/corr/abs-1811-08888, allen2018convergence, li2020rethinking}.
Under this premise, we tackle both GD and SGD in an unified framework and build on \cite{hayou2019mean} to model training of a linear model using a Stochastic Differential Equation in function space.
 We show that, as also hypothesized by \cite{mu2020gradients}, linearization can provide an accurate approximation of fine-tuning dynamics and therefore can be used for training time prediction.
\begin{figure}
    \centering
    \begin{subfigure}[t]{.49\linewidth}
        \centering
        \includegraphics[width=5cm]{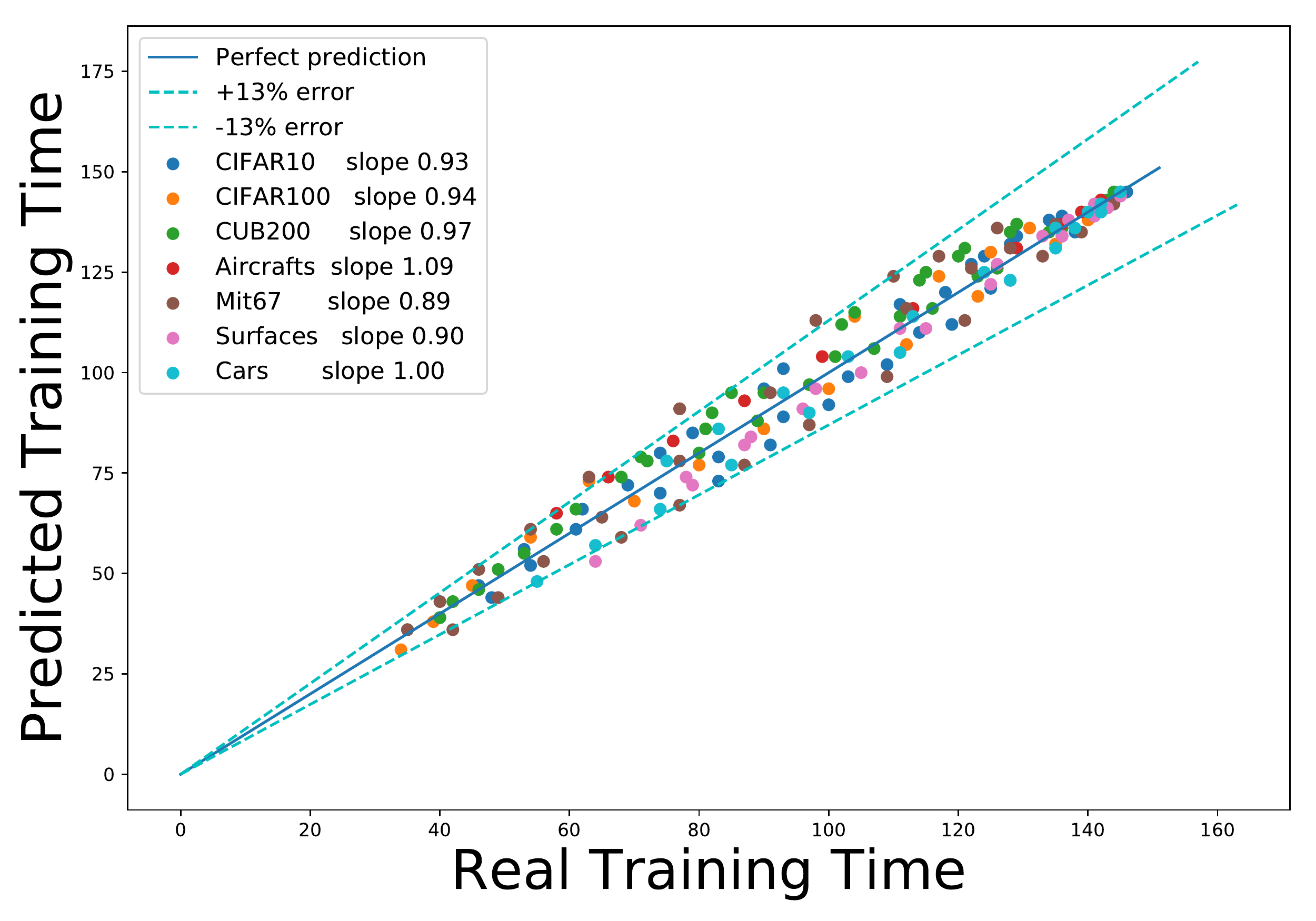}
        \caption{Training with Gradient Descent.}
    \end{subfigure}
    \begin{subfigure}[t]{.49\linewidth}
        \centering
        \includegraphics[width=5cm]{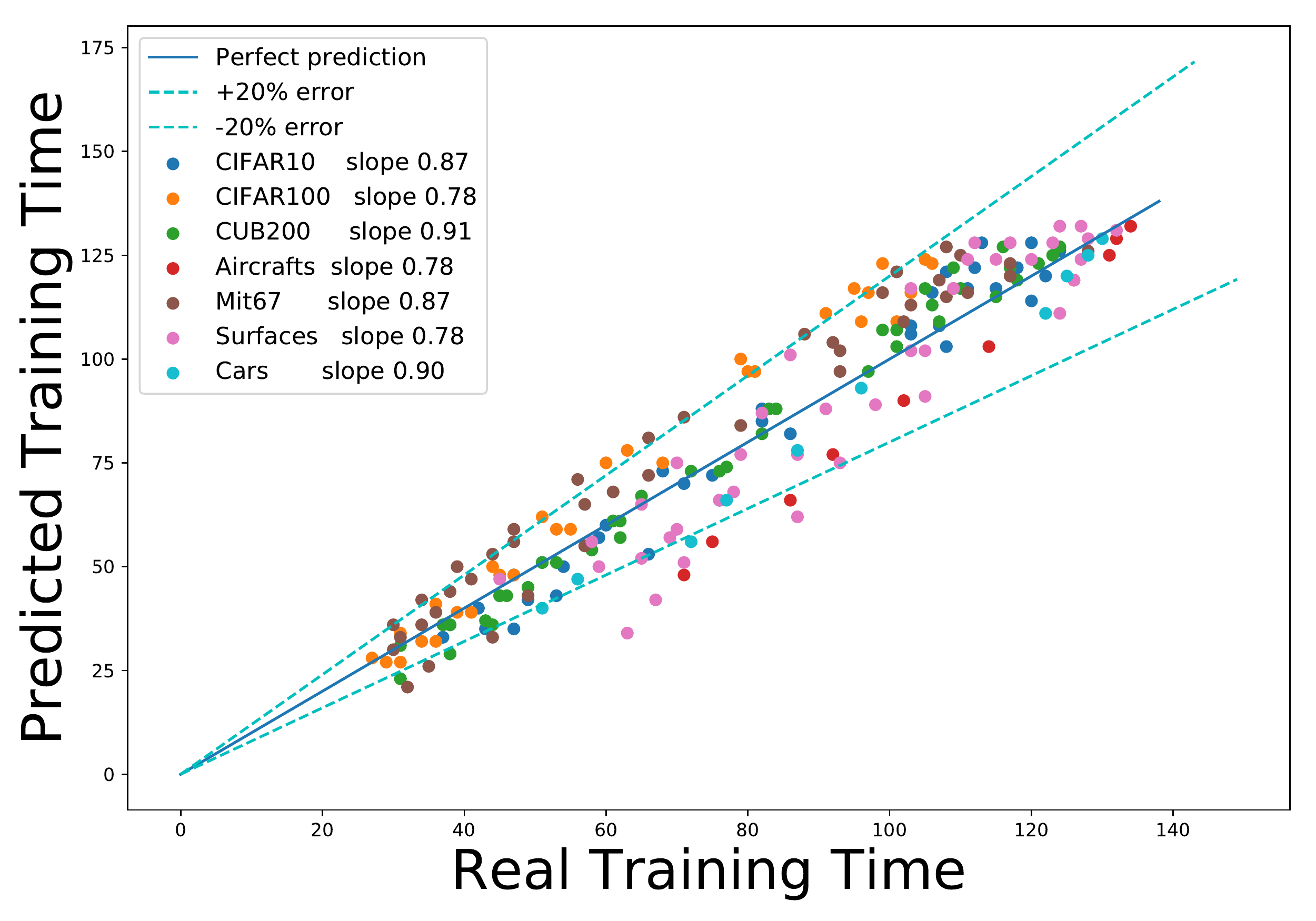}
        \caption{Training with SGD.}
    \end{subfigure}
    \caption{\textbf{Training time prediction (\# iterations) for several fine-tuning tasks.}
    Scatter plots of the predicted time vs the actual training time when fine-tuning a ResNet-18 pre-trained on ImageNet on several tasks. Each task is obtained by randomly sampling a subset of five classes with 150 images (when possible) each from one popular dataset  with different hyperparameters (batch size, learning rate). The closer the scatter plots to the bisector the better the TT estimate. Our prediction is \textbf{(a)} within 13\% of the real training time 95\% of the times when using GD and \textbf{(b)} within 20\% of the real training time when using SGD.
    }
    \label{fig:fine-tuning-time-prediction}
\end{figure}

\section{Predicting training time}
\label{sec:sde}
In this section we look at how to efficiently approximate the training time of a DNN without actual training. By \emph{Training Time} (TT) we mean the number of optimization steps -- of either Gradient Descent (GD) or Stochastic Gradient Descent (SGD) -- needed to bring the loss on the training set below a certain threshold.

\newcommand{\flt}{f^\lin_t}
\newcommand{\utext}[2]{\underbrace{#1}_{\text{#2}}}

We start by introducing our main tool. Let  $f_w(x)$ denote the output of the network, where $w$ denotes the weights of the network and $x \in \mathbb{R}^d$ denotes its input (e.g., an image). Let $w_0$ be the weight configuration after pre-training. We assume that when fine-tuning a pre-trained network the solution remains close to pre-trained weights $w_0$ \cite{mu2020gradients, pmlr-v97-du19c, DBLP:journals/corr/abs-1811-08888, allen2018convergence}.
Under this assumption -- which we discuss further in \Cref{sec:assumptions} -- we can faithfully approximate the network with its Taylor expansion around $w_0$ \cite{lee2019wide}. Let $w_t$ be the fine-tuned weights at time $t$. Using big-O notation and $f_t \equiv f_{w_t}$, we have:
\[
f_t(x) = f_0(x) + \nabla_w f_0(x)|_{w=w_0} (w_t - w_0) + O(\|w_t-w_0\|^2)
\]
We now want to use this approximation to characterize the training dynamics of the network during fine-tuning to estimate TT. In such theoretical analyses \cite{jacot2018neural, lee2019wide, arora2019fine} it is common to assume that the network is trained with Gradient Descent (GD) rather than Stochastic Gradient Descent, and in the limit of a small learning rate. In this limit, the dynamics are approximated by the gradient flow differential equation $\dot{w}_t = - \eta \nabla_{w_t} \L$
\cite{jacot2018neural, lee2019wide} where $\eta$ denotes the learning rate and $\L(w)$ denotes the loss function
$
\L(w) = \sum_{i=1}^N \ell(y_i, f_w(x_i)).
$,
where $\ell$ is the per-sample loss function (e.g. Cross-Entropy).
This approach however has two main drawbacks. First, it does not properly approximate  Stochastic Gradient Descent, as it ignores the effect of the gradient noise on the dynamics, which affects both training time and generalization.
Second, the differential equation involves the weights of the model, which live in a very high dimensional space thus making finding numerical solutions to the equation not tractable.

To address both problems, building on top of \cite{hayou2019mean} in the Supplementary we prove the following result.
\begin{prop}
\label{prop:activations-sde}
In the limit of small learning rate $\eta$, the \textit{output on the training set} of a linearized network $f_t^{lin}$ trained with SGD evolves according to the following Stochastic Differential Equation (SDE):
\begin{equation}
    \label{eq:activations-sde}
    df^\lin_t(\X) = \utext{- \eta \NTK \nabla_{f^\lin_t(\X)} \L_t\,  dt}{deterministic part} + \utext{\frac{\eta}{\sqrt{|B|}}  \nabla_{w} f^\lin_0(\X) \Sigma^{\frac{1}{2}}(f^\lin_t(\X)) dn}{stochastic part},
\end{equation}

where $\mathcal{X}$ is the set of training images, $|B|$ the batch-size and $dn$ is a $D$-dimensional Brownian motion. We have defined the Gram gradients matrix  $\NTK$ \cite{jacot2018neural, shawe2005eigenspectrum} (i.e., the empirical Neural Tangent Kernel matrix) and the covariance matrix $\Sigma$ of the gradients as follows:
\begin{align}
\NTK &:= \nabla_w f_0(\X) \nabla_w f_0(\X)^T, \label{eq:ntk-definition} \\
\Sigma(f^\lin_t(\X)) &:=  \E \big[ (g_i \nabla_{\flt(x_i)} \L) \otimes (g_i \nabla_{\flt(x_i)} \L) \big] - \E\big[g_i \nabla_{\flt(x_i)} \L\big] \otimes \E\big[g_i \nabla_{\flt(x_i)} \L\big]. \label{eq:varying-covariance}
\end{align}
where $g_i \equiv \nabla_w f_0(x_i)$. Note both $\Theta$ and $\Sigma$ only require gradients w.r.t. parameters computed at initialization.
\end{prop}
The first term of \cref{eq:activations-sde} is an ordinary differential equation (ODE) describing the deterministic part of the optimization, while the second stochastic term accounts for the noise. In \Cref{fig:effective-learning-rates} (left) we show the qualitative different behaviour of the solution to the deterministic part of \cref{eq:activations-sde} and the complete SDE \cref{eq:activations-sde}.
While several related results are known in the literature for the dynamics of the network in weight space \cite{DBLP:journals/corr/abs-1710-11029}, note that \cref{eq:activations-sde} completely characterizes the training dynamics of the linearized model by looking at the evolution of the output $f_t^{lin}(\X)$ of the model on the training samples -- a $N \times C$-dimensional vector -- rather than looking at the evolution of the weights $w_t$ -- a $D$-dimensional vector.
When the number of data points is much smaller than the number of weights (which are in the order of millions for ResNets), this can result in a drastic dimensionality reduction, which allows easy estimation of the solution to \cref{eq:activations-sde}. Solving \cref{eq:activations-sde} still comes with some challenges, particularly in computing  $\Theta$ efficiently on large datasets and architectures. We tackle these in \Cref{sec:computing-TT}. Before that, we take a look at how different hyper-parameters and different pre-trainings affect the training time of a DNN on a given task.

\subsection{Effect of hyper-parameters on training time}
\label{sec:effect-of-hyperparameters}
\begin{figure}
    \centering
    \includegraphics[height=3.5cm]{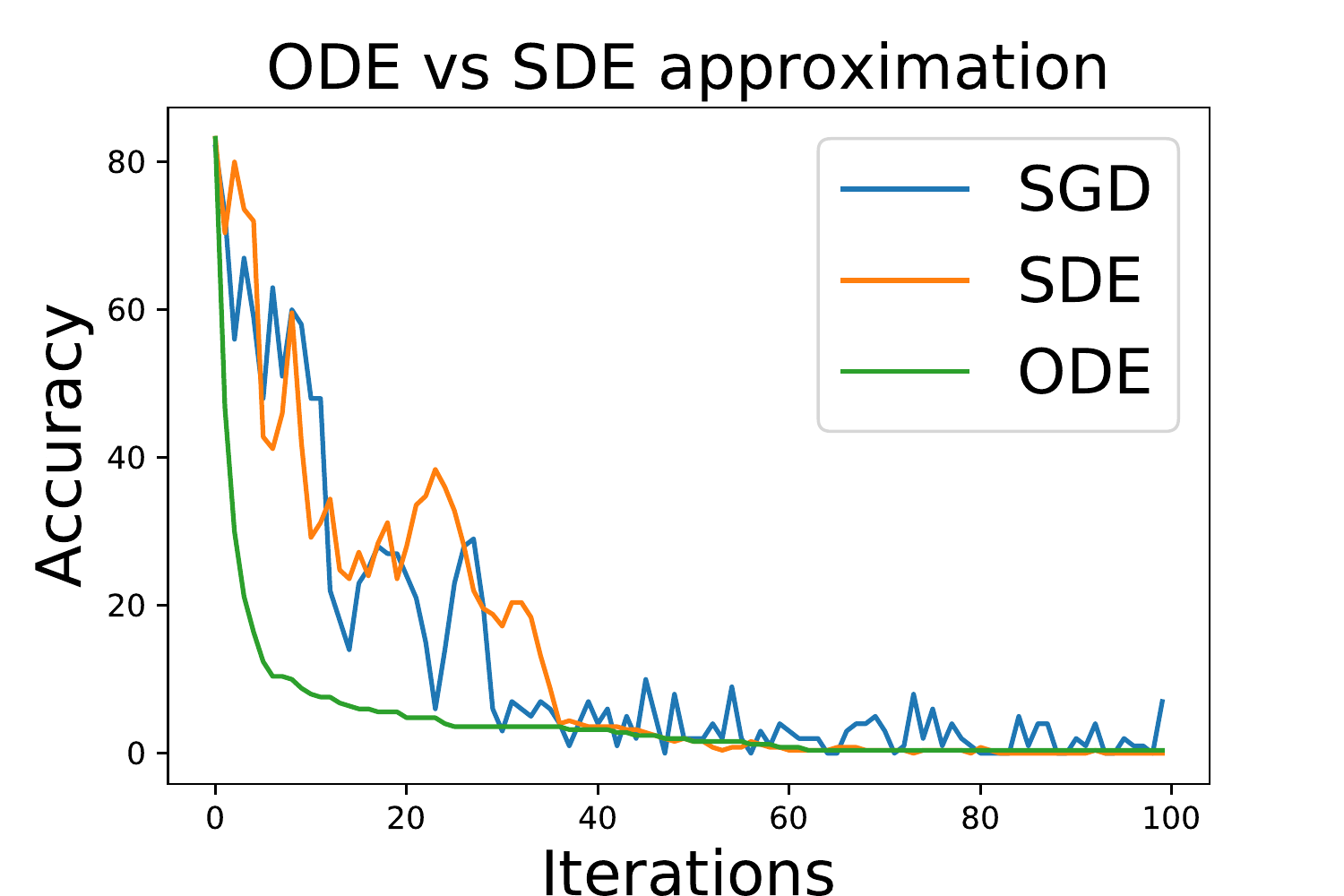}\hspace{.5cm}
    \includegraphics[height=3.5cm]{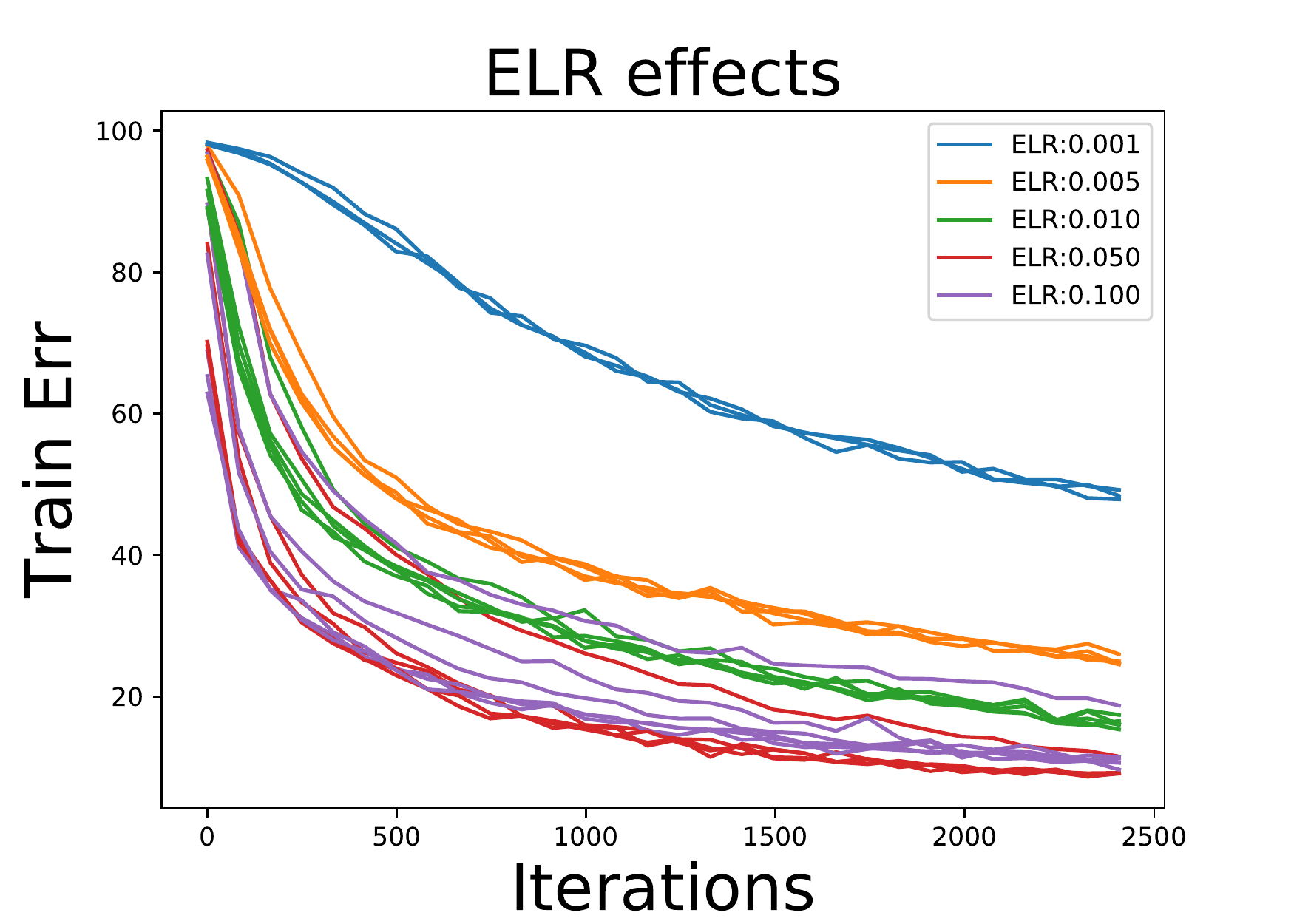}
    \caption{\textbf{(Left) ODE vs. SDE.} ODE approximation may not be well suited to describe the actual non-linear SGD dynamics (high learning rates regime).
    \textbf{(Right) Fine-tuning with the same ELR have similar curves}. We fine-tune an ImageNet pre-trained network on MIT-67 with different combinations of learning rates and momentum coefficients. We note that as long as the effective learning rate is the same, the loss curves are also similar.}
    \label{fig:effective-learning-rates}
\end{figure}

\textbf{Effective learning rate.} From \Cref{prop:activations-sde} we can gauge how hyper-parameters will affect the optimization process of the linearized model and, by proxy, of the original model it approximates. One thing that should be noted is that \Cref{prop:activations-sde} assumes the network is trained with momentum $m=0$. Using a non-zero momentum leads to a second order differential equation in weight space, that is not captured by \Cref{prop:activations-sde}. We can however, introduce heuristics to handle the effect of momentum: Smith et al. \cite{Smith2018ABP} note that the momentum acts on the stochastic part shrinking it by a factor $\sqrt{1/(1-m)}$. Meanwhile, under the assumptions we used in \Cref{prop:activations-sde} (small learning rate), we can show (see Supplementary Material) the main effect of momentum on the deterministic part is to re-scale the learning rates by a factor $1/(1-m)$.
Given these results, we define the effective learning rate (ELR) $\hat{\eta} = \eta/(1-m)$ and claim that, in first approximation, we can simulate the effect of momentum by using $\hat{\eta}$ instead of $\eta$ in \cref{eq:activations-sde}. In particular, models with different learning rates and momentum coefficients will have similar (up to noise) dynamics (and hence training time) as long as the effective learning rate $\hat{\eta}$ remains the same. In \Cref{fig:effective-learning-rates} we show empirically that indeed same effective learning rate implies similar loss curve. That similar effective learning rate gives similar test performance has also been observed in \cite{li2020rethinking, Smith2018ABP}.

\textbf{Batch size.} The batch size appears only in the stochastic part of the equation, its main effect is to decrease the scale of the SDE noise term. In particular, when the batch size goes to infinity $|B| \to \infty$ we recover the deterministic gradient flow also studied by \cite{lee2019wide}. Note that we need the batch size $|B|$ to go to infinity, rather than being as large as the dataset since we assumed random batch sampling with replacement. If we assume extraction without replacement the stochasticity is annihilated as soon as $|B|=N$ (see \cite{DBLP:journals/corr/abs-1710-11029} for a more in depth discussion).

\subsection{Effect of pre-training on training time}
\label{sec:pre-training}
\begin{figure}
    \centering
       \begin{subfigure}[t]{.49\linewidth}
            \includegraphics[width=6cm,height=3cm]{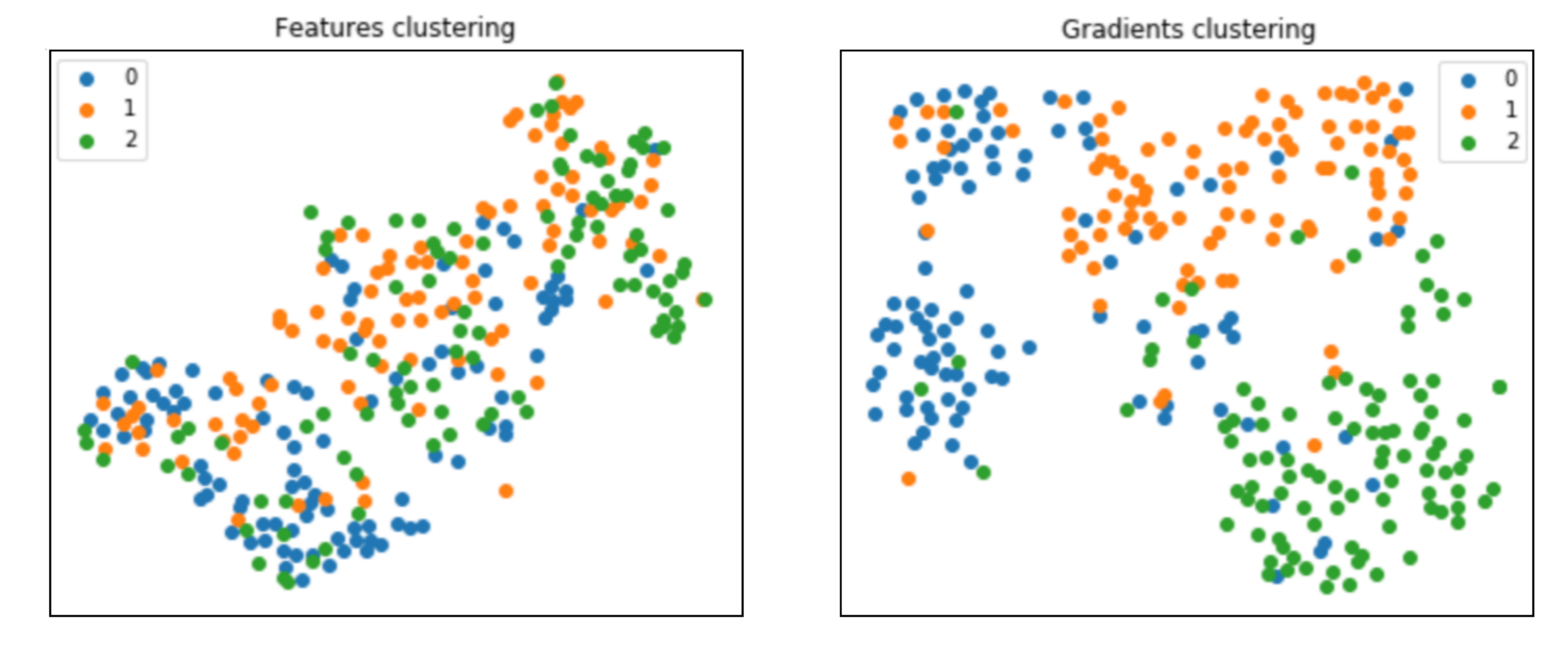}
            \caption{Features and Gradients clustering.}
        \end{subfigure}
        \begin{subfigure}[t]{.49\linewidth}
            \includegraphics[width=6cm,height=2.95cm]{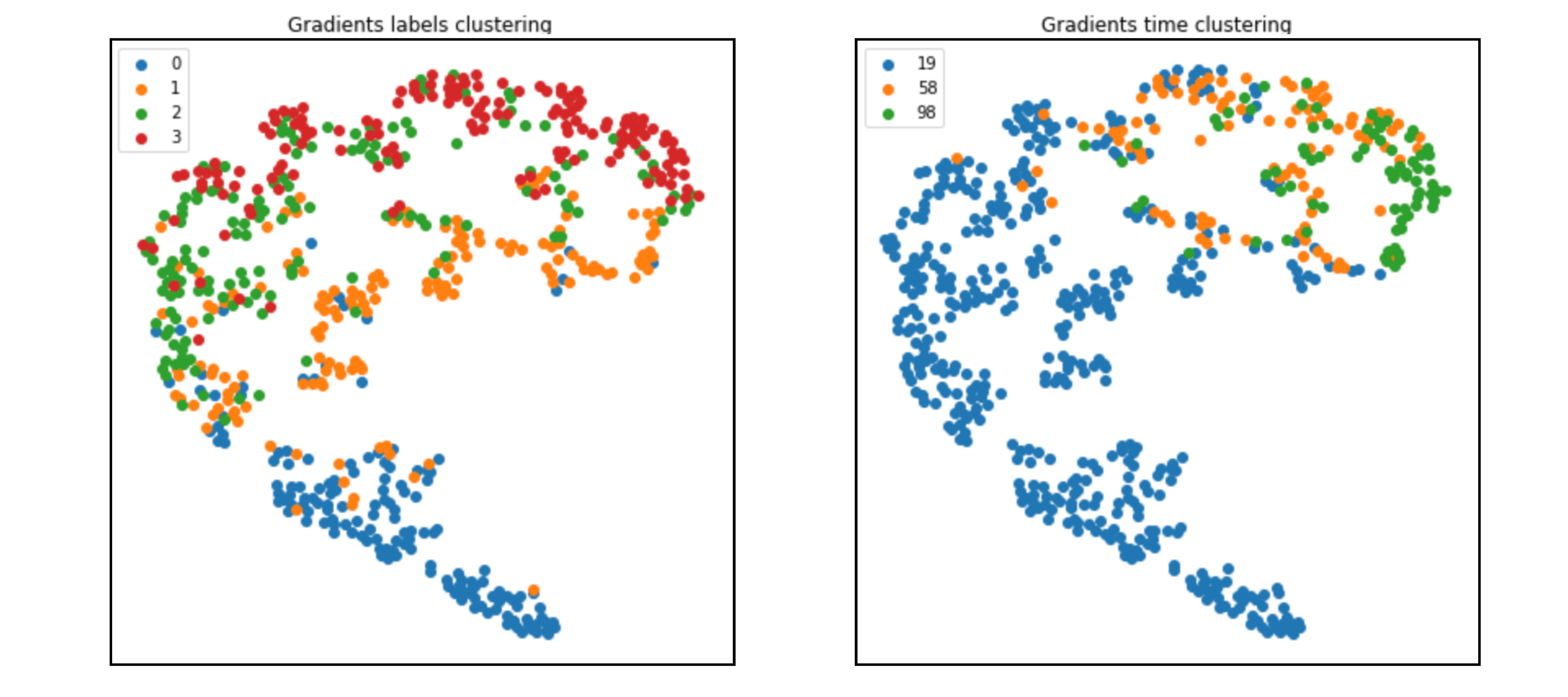}
            \caption{Trajectory clustering.}
        \end{subfigure}
        \caption{\textbf{Are gradients good descriptors to cluster data by semantics and training time?}
        (a) \textbf{Features vs Gradients clustering.}
        \textbf{(Right)} t-SNE plot of the first five principal components of the gradients of each sample in a subset of CIFAR-10 with 3 classes. Colors correspond to the sample class. We observe that the first 5 principal components are enough to separate the data by class. By \Cref{prop:mse-training-time} this implies faster training time.
        (\textbf{Left}) In the same setting as before, t-SNE plot of the features using the first 5 components of PCA. We observe that gradients separate the classes better than the features.
        \textbf{(b) t-SNE on predicted trajectories} To see if gradients are good descriptors of both semantics and training time we use gradients to predict linearized trajectories: we cluster the trajectories using t-SNE and we color each point by
        \textbf{(left)} class and
        \textbf{(right)} training time. We observe that: clusters split trajectories according both to labels \textbf{(left)} and training time  \textbf{(right)}. Interestingly inside each class there are clusters of points that may converge at different speed.
        }
        \label{fig:trajectories-clustering}
\end{figure}

We now use the SDE in \cref{eq:activations-sde} to analyze how the combination of different pre-trainings of the model -- that is, different $w_0$'s -- and different tasks affect the training time. In particular,
we show that a necessary condition for fast convergence is that the gradients after pre-training cluster well with respect to the labels.
We conduct this analysis for a binary classification task with $y_i = \pm 1$, but the extension is straightforward for multi-class classification, under the simplifying assumptions that we are operating in the limit of large batch size (GD) so that only the deterministic part of \cref{eq:activations-sde} remains.
Under these assumptions, \cref{eq:activations-sde} can be solved analitically and the loss of the linearized model at time $t$ can be written in closed form as (see Supplementary Material):
\begin{align}
\boxed{L_t = (\Y - f_0(\X))^T e^{- 2 \eta \NTK t}(\Y - f_0(\X))}
\label{eq:mse-loss-dynamics}
\end{align}

The following characterization can easily be obtained using an eigen-decomposition of the matrix $\Theta$.
\begin{prop}
\label{prop:mse-training-time}
Let $S = \nabla_w f_w(\X)^T \nabla_w f_w(\X)$ be the second moment matrix of the gradients and let $S = U \Sigma U^T$ be the uncentered PCA of the gradients, where $\Sigma = \operatorname{diag}({\lambda_1, \ldots, \lambda_n, 0, \ldots, 0})$ is a $D \times D$ diagonal matrix, $n\leq \min (N, D)$ is the rank of $S$ and $\lambda_i$ are the eigenvalues sorted in descending order. Then we have
\begin{equation}
\label{eq:dynamics-svd}
L_t = \sum_{k=1}^D e^{-2\eta\lambda_k t} (\dy \cdot \v_k)^2,
\end{equation}
where $\lambda_k \v_k = (g_i \cdot \u_k)_{i=1}^N$ is the $N$-dimensional vector containing the value of the $k$-th principal component of gradients $g_i$ and $\dy:= \mathcal{Y} - f_0(\mathcal{X})$.
\end{prop}
\textbf{Training speed and gradient clustering.} We can give the following intuitive interpretation: consider the gradient vector $g_i$ as a representation of the sample $x_i$. If the first principal components of $g_i$ are sufficient to separate the classes (i.e., cluster them), then convergence is faster (see \Cref{fig:trajectories-clustering}). Conversely, if we need to use the higher components (associated to small $\lambda_k$) to separate the data, then convergence will be exponentially slower.
Arora et al. \cite{arora2019fine} also use the eigen-decomposition of $\Theta$ to explain the slower convergence observed for a randomly initialized two-layer network trained with random labels. This is straightforward since the projection of a random vector will be uniform on all eigenvectors, rather than concentrated on the first few, leading to slower convergence. However, we note that the exponential dynamics predicted by \cite{arora2019fine} do not hold for more general networks trained from scratch \cite{DBLP:conf/iclr/ZhangBHRV17} (see \Cref{sec:assumptions}).
In particular, \cref{eq:dynamics-svd} mandates that the loss curve is always convex (it is sum of convex functions), which may not be the case for deep networks trained from scratch.

\section{Efficient numerical estimation of training time} \label{sec:computing-TT}
\begin{figure}
    \centering
    \includegraphics[height=2.5cm]{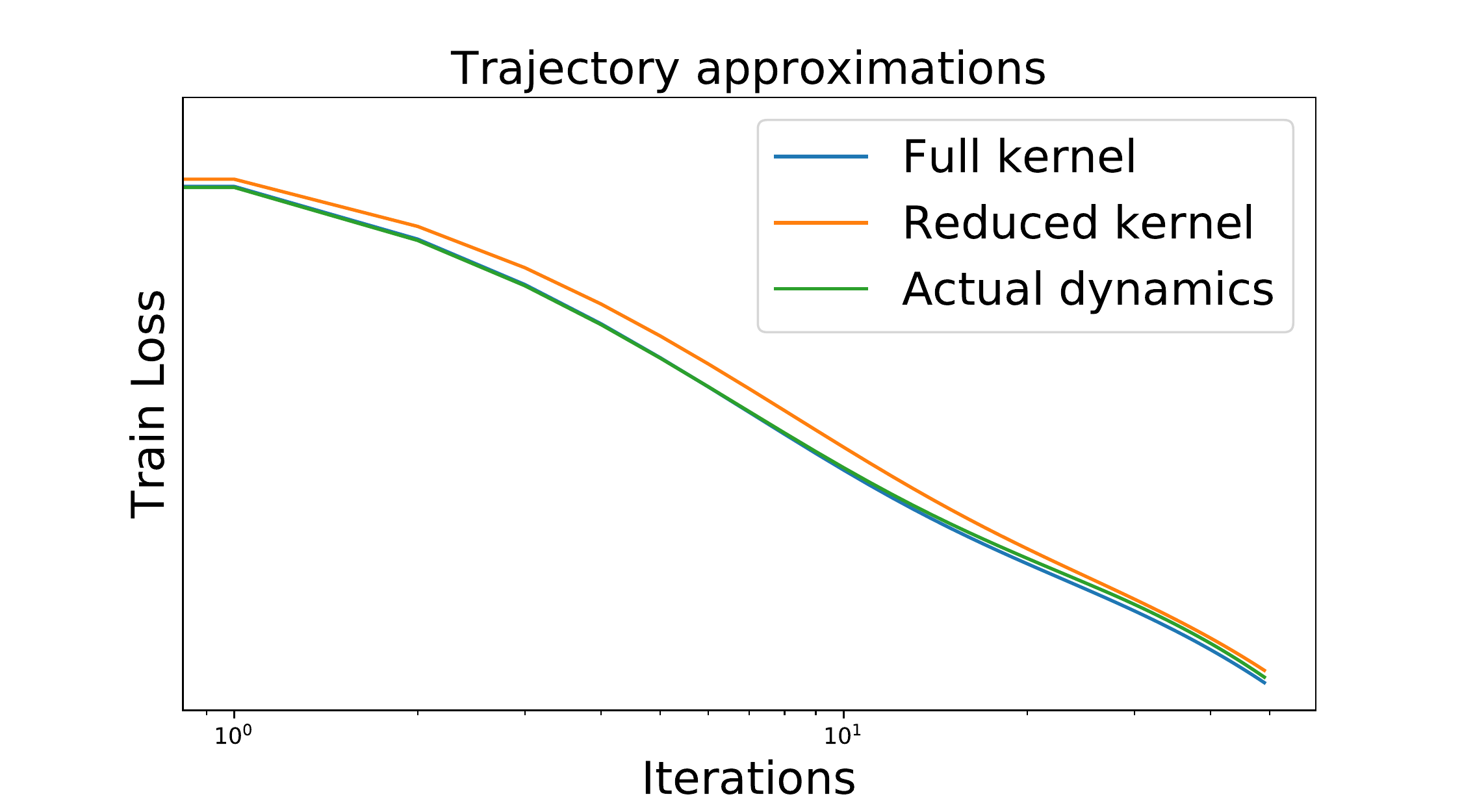}
    \includegraphics[height=2.5cm]{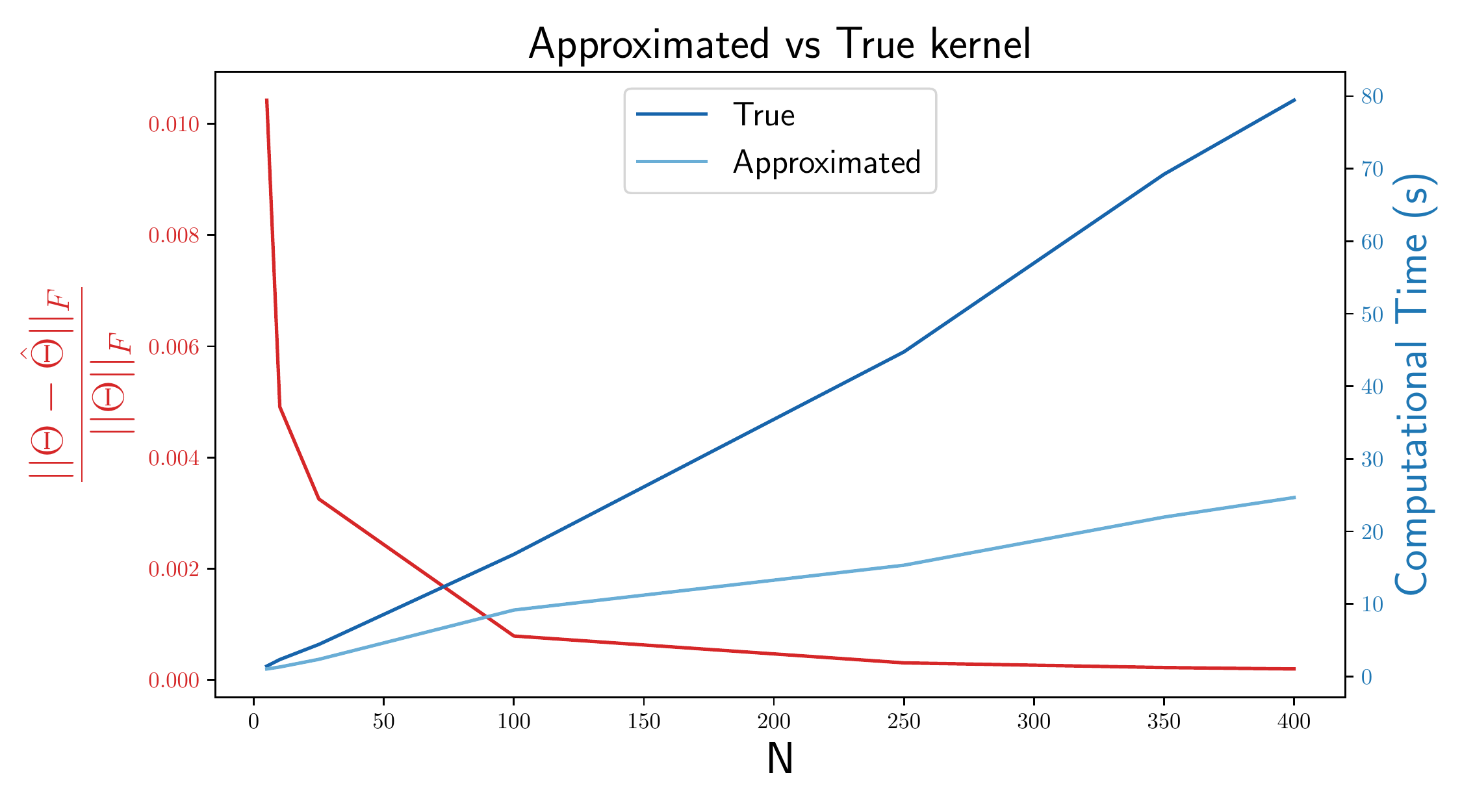}
    \includegraphics[height=2.5cm]{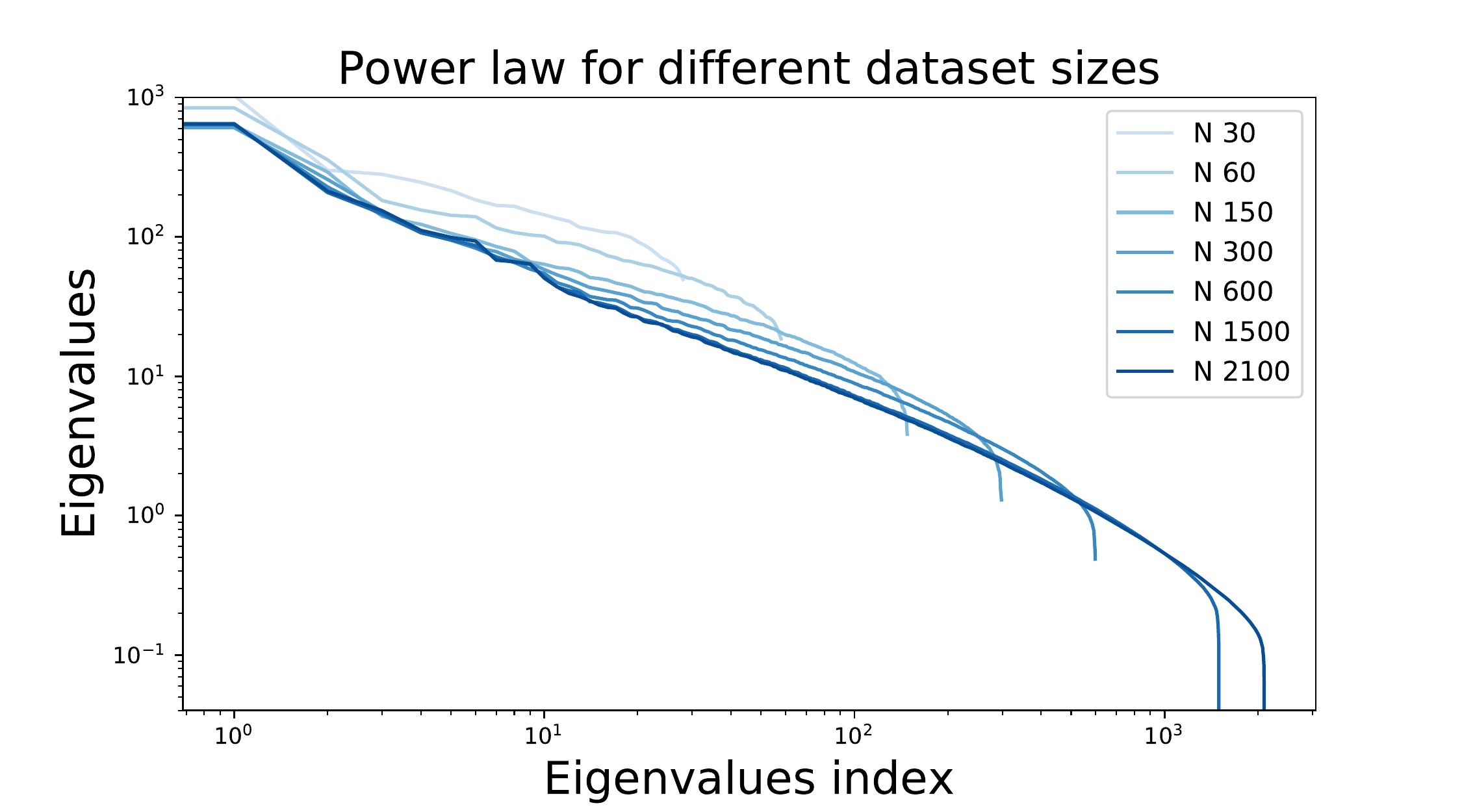}
    \caption{
        \textbf{(Left)} Actual fine-tuning of a DNN with GD compared to the numerical solution of \cref{eq:activations-sde} and the solution using an approximated $\Theta$. The approximated $\NTK$ can faithfully describe fine-tuning dynamics while being twice as fast to compute and 100 times smaller to be stored.
        \textbf{(Center)} Relative difference in Frobenius norm of the real and approximated $\Theta$ as the dataset size varies (red), and their computational time (blue).
        \textbf{Right}: Eigen-spectrum of $\Theta$ computed on subsets of MIT-67 of increasing size.  Note the convergence to a common power law (i.e., a line in log-log scale).
    }
    \label{fig:NTK-approximation}
\end{figure}

In \Cref{prop:mse-training-time} we have shown a closed form solution to the SDE in \cref{eq:activations-sde} in the limit of large batch size, and for the MSE loss. Unfortunately, in general \cref{eq:activations-sde} does not have a closed form expression
when using the cross-entropy loss \cite{lee2019wide}. A numerical solution is however possible, enabled by the fact that we describe the network training in function space, which is much smaller than weight space for over-parametrized models.
The main computational cost is to create the matrix $\NTK$ in \cref{eq:activations-sde} -- which has cost $O(D C^2 N^2)$ -- and to compute the noise in the stochastic term.
Here we show how to reduce the cost of $\NTK$ to $O(D_0 C^2 N^2)$ for $D_0 \ll D$ using a random projection approximation. Then, we propose a fast approximation for the stochastic part.
Finally, we describe how to reduce the cost in $N$ by using only a subset $N' < N$ of samples to predict training time.

\textbf{Random projection.} \label{sec:grad-rand-proj}
To keep the notation uncluttered, here we assume  w.l.o.g. $C=1$. In this case the matrix $\Theta$ contains $N^2$ pairwise dot-products of the gradients (a $D$-dimensional vector) for each of the $N$ training samples (see eq.~\ref{eq:ntk-definition}). Since $D$ can be very large (in the order of millions) storing and multiplying all gradients can be expensive as $N$ grows. Hence, we look at a dimensionality reduction technique. The optimal dimensionality reduction that preserves the dot-product is obtained by projecting on the first principal components of SVD, which however are themselves expensive to obtain. A simpler technique is to project the gradients on a set of $D'$ standard Gaussian random vectors: it is known that such random projections preserve (in expectation) pairwise product \cite{bingham2001random, 10.1016/S0022-0000(03)00025-4} between vectors, and hence allow us to reconstruct the Gram matrix while storing only $D'$-dimensional vector, with $D' \ll D$. We further increase computational efficiency using multinomial random vectors \{-1,0,+1\} as proposed in \cite{10.1016/S0022-0000(03)00025-4} which further reduce the computational cost by avoiding floating point multiplications. In \Cref{fig:NTK-approximation} we show that the entries of $\Theta$ and its spectrum are well approximated using this method, while the computational time becomes much smaller.

\textbf{Computing the noise.} The noise covariance matrix $\Sigma$ is a $D\times D$-matrix that changes over time. Both computing it at each step and storing it is prohibitive. Estimating $\Sigma$ correctly is important to describe the dynamics of SGD \cite{chaudhari2018stochastic}, however we claim that a simple approximation may suffice to describe the simpler dynamic in function space. We approximate $ \nabla_{w} f^\lin_0(\X) \Sigma^{1/2}$ approximating $\Sigma$ with its diagonal (so that the we only need to store a $D$-dimensional vector). Rather than computing the whole $\Sigma$ at each step, we estimate the value of the diagonal at the beginning of the training. Then, by exploiting \cref{eq:varying-covariance}, we see that the only change to $\Sigma$ is due to $\nabla_{\flt} \L$, whose norm decreases over time. Therefore we use the  easy-to-compute $\nabla_{\flt} \L$ to re-scale our initial estimate of $\Sigma$.

\textbf{Larger datasets.} In the MSE case from \cref{eq:mse-loss-dynamics}, knowing the eigenvalues $\lambda_k$ and the corresponding residual projections $p_k = (\dy \cdot \v_k)^2$ we can predict in closed form the whole training curve. Is it possible to predict $\lambda_k$ and $p_k$ using only a subset of the dataset? It is known \cite{shawe2005eigenspectrum} that the eigenvalues of the Gram matrix of Gaussian data follow a power-law distribution of the form $\lambda_k = c k^{-s}$. Moreover, by standard concentration argument, one can prove that the eigenvalues should converge to a given limit as the number of datapoints increases. We verify that a similar power-law and convergence result also holds for real data (see \Cref{fig:NTK-approximation}). Exploiting this result, we can estimate $c$ and $s$ from the spectrum computed on a subset of the data, and then predict the remaining eigenvalues. A similar argument holds for the projections $p_k$, which also follow a power-law (albeit with slower convergence). We describe the complete estimation in the Supplementary Material.

\section{Results}
We now empirically validate the accuracy of \cref{prop:activations-sde} in approximating the loss curve of an actual deep neural network fine-tuned on a large scale dataset. We also validate the goodness of the numerical approximations described in \Cref{sec:computing-TT}. Due to the lack of a standard and well established benchmark to test Training Time estimation algorithms we developed one with the main goal to closely resemble fine-tuning common practice for a wide spectrum of different tasks.

\textbf{Experimental setup.}
We define training time as the first time the (smoothed) loss is below a given threshold. However, since different datasets converge at different speeds, the same threshold can be too high (it is hit immediately) for some datasets, and too low for others (it may take hundreds of epochs to be reached). To solve this, and have cleaner readings, we define a `normalized' threshold as follows: we fix the total number of fine-tuning steps $T$, and measure instead the first time the loss is within $\epsilon$ from the final value at time $T$. This measure takes into account the `asymptotic' loss reached by the DNN within the computational budget (which may not be close to zero if the budget is low), and naturally adapts the threshold to the difficulty of the dataset.
We compute both the real loss curve and the predicted training curve using \Cref{prop:activations-sde} and compare the $\epsilon$-training-time measured on both. We report the \textit{absolute prediction error}, that is $|t_\text{predicted} - t_\text{real}|$.
For all the experiments we extract 5 random classes from each dataset (\Cref{tab:TT-error}) and sample 150 images (or the maximum available for the specific dataset). Then we fine-tuned ResNet18/34 using either GD or SGD.

\textbf{Accuracy of the prediction.} In \Cref{fig:fine-tuning-time-prediction} we show TT estimates errors (for different $\epsilon \in \{1,...,40\}$) under a plethora of different conditions ranging from different learning rates, batch sizes, datasets and optimization methods. For all the experiments we choose a multi-class classification problem with Cross Entropy (CE) Loss unless specified otherwise, and fixed computational budget of  $T=150$ steps both for GD and SGD. We note that our estimates are consistently within respectively a 13\% and 20\% relative error around the actual training time 95\% of the times.

In \Cref{tab:TT-error} we describe the sensitivity of our estimates to different thresholds $\epsilon$ both when our assumptions do and do not hold (high and low learning rates regimes).
Note that a larger threshold $\epsilon$ is hit during  the initial convergence phase of the network, when a small number of iterations corresponds a large change in the loss. Correspondingly, the hitting time can be measured more accurately and our errors are lower. A smaller $\epsilon$ depends more on correct prediction of the slower asymptotic phase, for which exact hitting time is more difficult to estimate.

\vspace{-0.8em}

\begin{table}[ht]
  \begin{varwidth}[b]{0.48\linewidth}
    \centering
    \tiny
    \begin{tabular}{|l||c|c||c|c||c|c|}
    \hline
     TT error (\# of steps) & \multicolumn{2}{c|}{\bf $\epsilon$ = 1\%} & \multicolumn{2}{c|}{\bf $\epsilon$ = 10\% }  & \multicolumn{2}{c|}{{\bf $\epsilon$= 40\%}} \\
     Lr       & low & high & low & high & low & high \\
    \hline
    Cars \cite{KrauseStarkDengFei-Fei_3DRR2013}     &           9 & 18 &  7 &  8 & 1 & 0 \\
    Surfaces \cite{bell15minc}                      &            6 &  13 &  6 &  7 & 6 & 3\\
    Mit67 \cite{conf/cvpr/QuattoniT09}              &            8 &  10 &  6 &  8 &  3 & 1 \\
    Aircrafts \cite{maji13fine-grained}             &            5 &  21 & 5 &  4 &  9 & 7 \\
    CUB200  \cite{WelinderEtal2010}                 &            6 &  6 &  5 &  8 &  1 & 1 \\
    CIFAR100 \cite{krizhevsky:2009}                 &           10 &  15 & 6 &  7 & 2 & 3 \\
    CIFAR10 \cite{krizhevsky:2009}                  &            9 &  14 & 8 &  9 & 3 & 3 \\
    \hline
    \end{tabular}
    \vspace{0.2cm}
    \caption{Training Time estimation error for CE loss using GD for $T=150$ epochs at different thresholds $\epsilon$. We compare TT estimates when ODE assumptions do and do not hold: high \{0.005\} and small LR \{0.001, 0.0001\}.
    }
    \label{tab:TT-error}
  \end{varwidth}%
  \hfill
  \begin{minipage}[b]{0.48\linewidth}
    \centering
        \includegraphics[width=5.8cm]{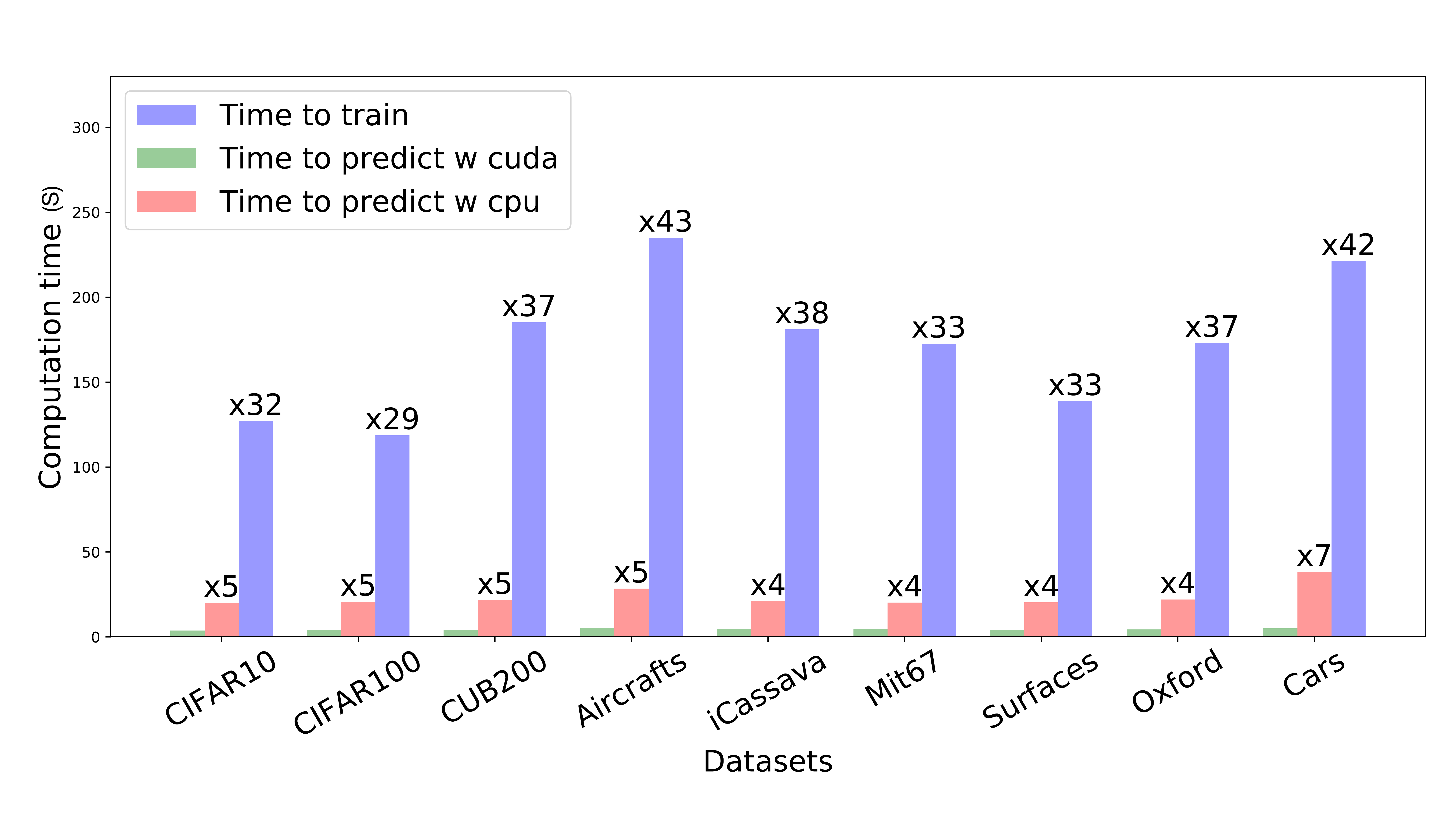}
         \captionof{figure}{Wall clock time (in seconds) to compute TT estimate vs fine-tuning running time.
        We run the methods described in \Cref{sec:grad-rand-proj} both on GPU and CPU. Training is done on GPU.
         }
        \label{fig:computational-cost}
  \end{minipage}
\end{table}

\vspace{-0.5em}

\textbf{Wall-clock run-time.} In \Cref{fig:computational-cost} we show the wall-clock runtime of our training time prediction method compared to the time to actually train the network for $T$ steps. Our method is 30-40 times faster. Moreover, we note that it can be run completely on CPU without a drastic drop in performance. This allows to cheaply estimate TT and allocate/manage resources even without access to a GPU.

\begin{figure}
    \centering
    \includegraphics[height=3cm]{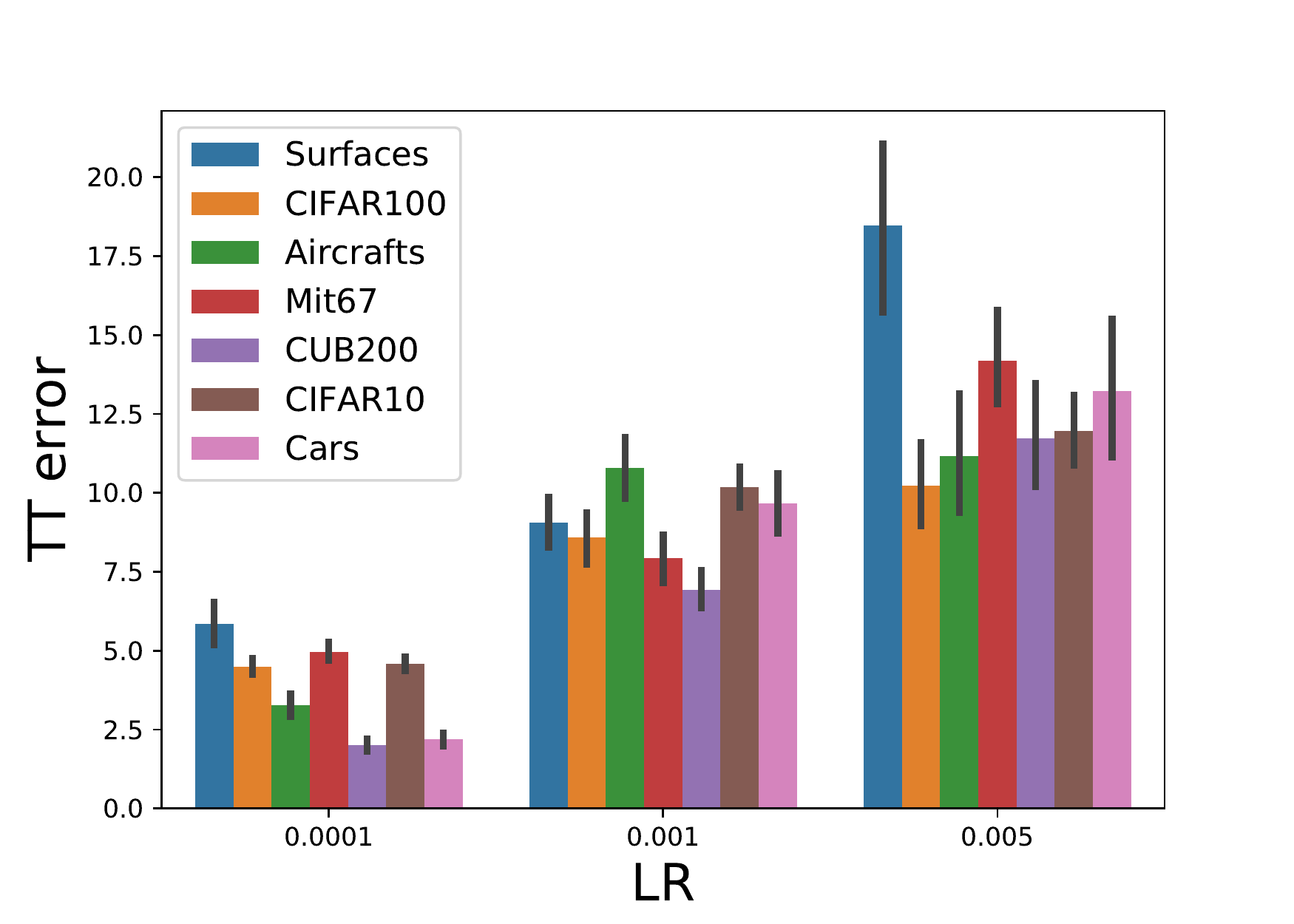}
    \includegraphics[height=3cm]{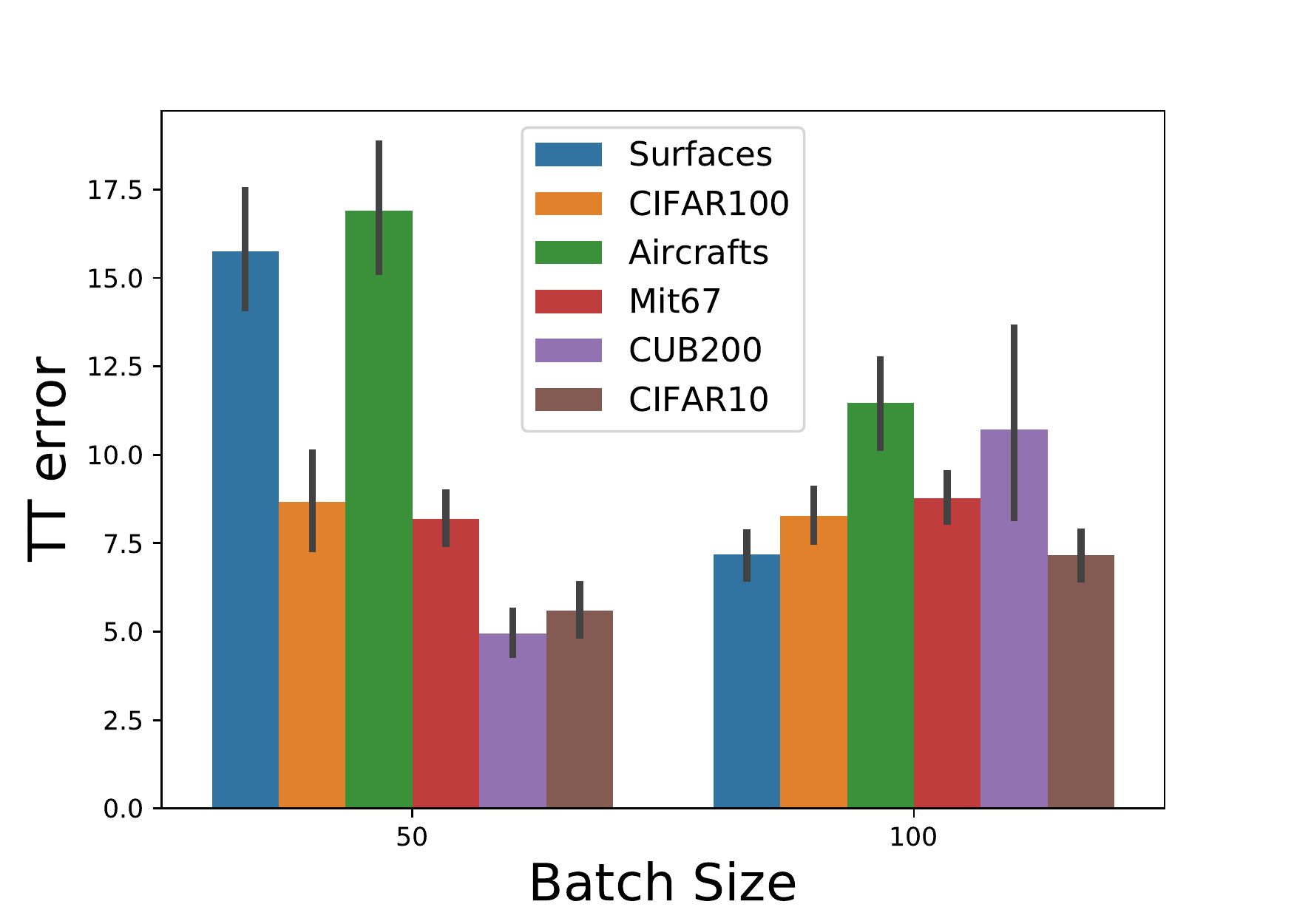}
    \includegraphics[height=3cm]{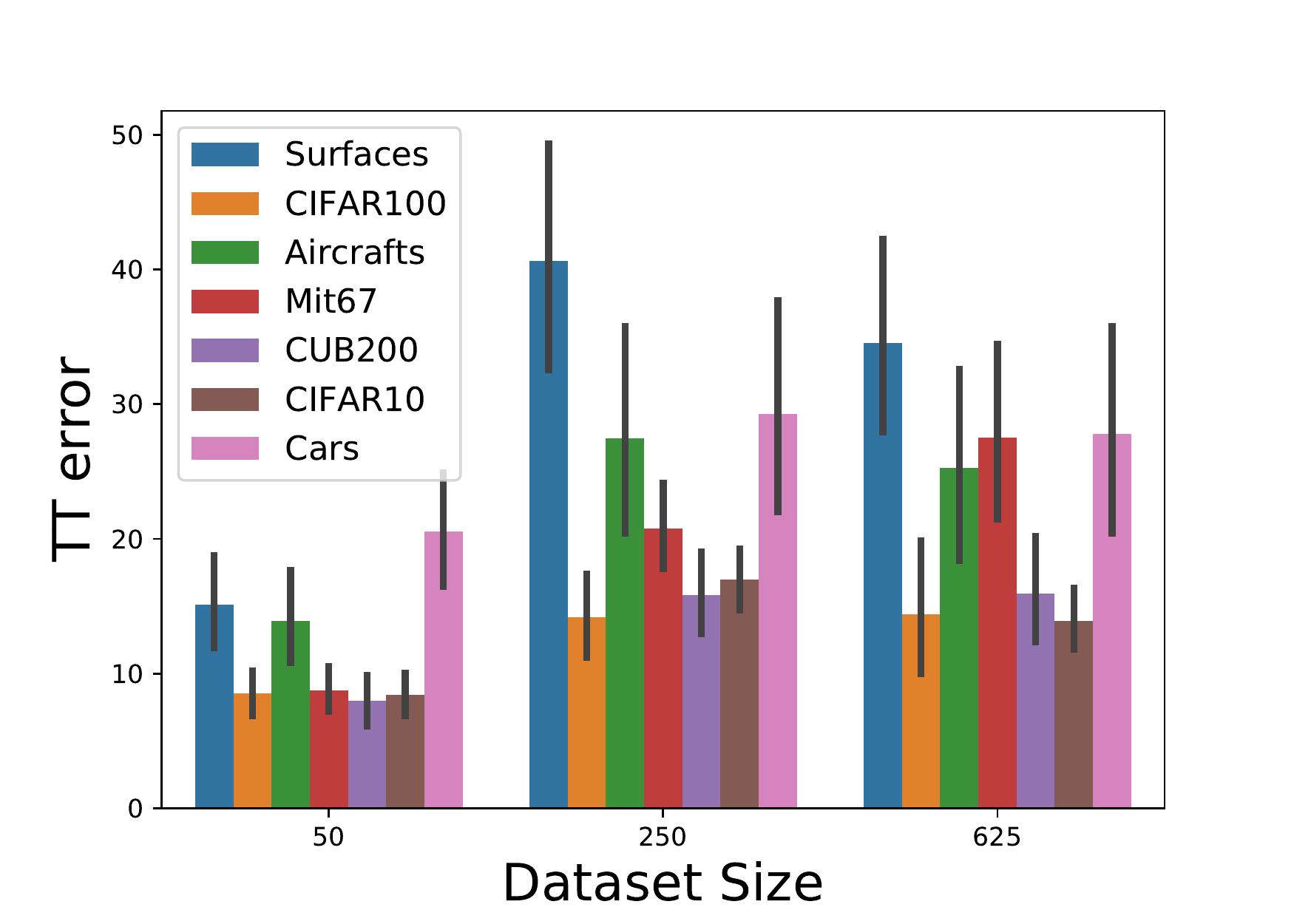}
    \caption{Average and 95\% confidence intervals of TT estimate error for:
        \textbf{Left}: GD using different learning rates.
        \textbf{Center}: SGD using different batch sizes.
        \textbf{Right}: SGD using different dataset sizes. The average is taken w.r.t. random classes with different number of samples: \{10, 50, 125\}
    }
    \label{fig:TT-error-HPs}
\end{figure}

\textbf{Effect of dataset distance.} We note that the average error for Surfaces (\Cref{fig:TT-error-HPs}) is uniformily higher than the other datasets. This may be due to the texture classification task being quite different from ImageNet, on which the network is pretrained. In this case we can expect that the linearization assumption is partially violated since the features must adjust more during fine-tuning.

\textbf{Effect of hyper-parameters on prediction accuracy.} We derived \Cref{prop:activations-sde} under several assumptions, importantly: small learning rate and $w_t$ close to $w_0$. In \Cref{fig:TT-error-HPs} (left) we show that indeed increasing the learning rate decreases the accuracy of our prediction, albeit the accuracy remains good even at larger learning rates. Fine-tuning on larger dataset makes the weights move farther away from the initialization $w_0$. In \Cref{fig:TT-error-HPs} (right) we show that this  slightly increases the prediction error.
Finally, we observe in \Cref{fig:TT-error-HPs} (center) that using a smaller batch size, which makes the stochastic part of \Cref{prop:activations-sde} larger also slightly increases the error.
This can be ascribed to the approximation of the noise term (\Cref{sec:grad-rand-proj}).
On the other hand, in \Cref{fig:effective-learning-rates} (right) we see that the effect of momentum on a fine-tuned network is very well captured by the effective learning rate (\Cref{sec:effect-of-hyperparameters}), as long as the learning rate is reasonably small, which is the case for fine-tuning. Hence the SDE approximation is robust to different values of the momentum. In general, we note that even when our assumptions are not fully met training time can still be approximated with only a slightly  higher  error.
This suggest that point-wise proximity of the training trajectory of linear and real models is not necessary as long as their behavior (decay-rate) is similar (see also  Supplementary Material).

\section{Discussion and conclusions}

\label{sec:assumptions}

We have shown that we can predict with a 13-20\% accuracy the time that it will take for a pre-trained network to reach a given loss, in only a small fraction of the time that it would require to actually train the model. We do this by studying the training dynamics of a linearized version of the model -- using the SDE in \cref{eq:activations-sde} -- which, being in the smaller function space compared to parameters space, can be solved numerically. We have also studied the dependency of training time from pre-training and hyper-parameters (\Cref{sec:effect-of-hyperparameters}), and how to make the computation feasible for larger datasets and architectures (\Cref{sec:computing-TT}).

While we do not necessarily expect a linear approximation around a random initialization to hold during training of a real (non wide) network, we exploit the fact that when using a pre-trained network the weights are more likely to remain close to initialization \cite{mu2020gradients}, improving the quality of the approximation. However, in the Supplementary Material we show that even when using a pre-trained network, the trajectories of the weights of linearized model and of the real model can differ substantially. On the other hand, we also show that the linearized model correctly predicts the \textit{outputs} (not the weights) of the real model throughout the training, which is enough to compute the loss. We hypothesise that this is the reason why \cref{eq:activations-sde} can accurately predict the training time using a linear approximation.

The procedure described so far can be considered as an open loop procedure meaning that,  since we are estimating training time before any fine-tuning step is performed, we are not gaining any feedback from the actual training.
How to perform training time prediction during the actual training, and use training feedback (e.g., gradients updates) to improve the prediction in real time, is an interesting future direction of research.

\bibliography{bibliography}
\bibliographystyle{plain}

\newpage

\appendix

\section*{Predicting Training Time Without Training: Supplementary Material}

In the Supplementary Material we give the pseudo-code for the training time prediction algorithm (\Cref{apx:algorithm}) together with implementation details, show additional results including prediction of training time using only a subset of samples, and comparison of real and predicted loss curves in a variety of conditions (\Cref{apx:additional-experiments}). Finally, we give proofs of all statements.

\section{Algorithm}
\label{apx:algorithm}

\begin{algorithm}[H]
\caption{Estimate the Training Time on a given target dataset and hyper-parameters.}
\begin{algorithmic}[1]
\SetAlgoLined
\STATE \textbf{Data:} {Number of steps $T$ to simulate, threshold $\epsilon$ to determine convergence, pre-trained weights $w_0$ of the model, a target dataset with images $\mathcal{X} = \{x_i\}_{i=1}^N$ and labels $\mathcal{Y}=\{y_i\}_{i=1}^N$, batch size $B$, learning rate $\eta$, momentum $m \in [0,1)$.}
\STATE \textbf{Result:} {An estimate $\hat{T}_\epsilon$ of the number of steps necessary to converge within $\epsilon$ to the final value ${T}_\epsilon:= \min \{t:|\mathcal{L}_t - \mathcal{L}_T| < \epsilon\}$.}
\STATE \textbf{Initialization:} Compute initial network predictions $f_0(\mathcal{X})$, estimate $\NTK$ using random projections (\Cref{sec:computing-TT}), compute the ELR $\tilde{\eta}=\eta/(1-m)$ to use in \cref{eq:activations-sde} instead of $\eta$\;
 \IF{ B = N }
 \STATE Get $f^\lin_t(\mathcal{X})$ solving the ODE in \cref{eq:activations-sde} (only the \textit{deterministic part}) for $T$ steps\;
 \ELSE
 \STATE Get $f^\lin_t(\mathcal{X})$ solving the SDE in \cref{eq:activations-sde} for $T$ steps (see approximation in \Cref{sec:computing-TT})\;
 \ENDIF
 \STATE Using $f^\lin_t(\mathcal{X})$ and $\mathcal{Y}$ compute linearized loss $\mathcal{L}^{lin}_t \quad \forall t \in \{1,...,T\}$
 \RETURN $\hat{T}_\epsilon:= \min \{t: |\mathcal{L}^{lin}_t - \mathcal{L}^{lin}_T|< \epsilon \}$\;
\end{algorithmic}
\end{algorithm}

We can compute the estimate on training time based also on the accuracy of the model: we straightforwardly modify the above algorithm and use the predictions $f_t^\lin(\X)$ to compute the error instead of the loss (e.g. \cref{fig:Comparing-Loss-Err-TT-predictions}).

We now briefly describe some implementations details regarding the numerical solution of ODE and SDE. Both of them can be solved by means of standard algorithms: in the ODE case we used LSODA (which is the default integrator in \textit{scipy.integrate.odeint}), in the SDE case we used Euler-Maruyama algorithm for Ito equations.

We observe removing batch normalization (preventing the statistics to be updated) and removing data augmentation improve linearization approximation both in the case of GD and SGD. Interestingly data augmentation only marginally alters the spectrum of the Gram matrix $\NTK$ and has little impact on the linearization approximation w.r.t. batch normalization. \cite{goldblum2019truth} observed similar effects but, differently from us, their analysis has been carried out using randomly initialized ResNets.

\section{Target datasets}

\begin{table}[!h]
\begin{center}
\setlength{\tabcolsep}{3pt}
\resizebox{\textwidth}{!}{
\begin{tabular}{|l||c|c|c|c||}
\hline
\bf{Dataset} & Number of images & Classes & Mean samples per class & Imbalance factor \\
\hline
\hline
cifar10 \cite{krizhevsky:2009}   & 50000	&10	&5000	&1 \\
cifar100 \cite{krizhevsky:2009}  & 50000	&100	&500	&1 \\
cub200 \cite{WelinderEtal2010}  &5994	&200	&29.97	&1.03 \\
fgvc-aircrafts \cite{maji13fine-grained}  &6667 & 100 & 66.67 & 1.02 \\
mit67 \cite{conf/cvpr/QuattoniT09}   & 5360	&67	&80	&1.08 \\
opensurfacesminc2500 \cite{bell15minc}           & 48875	&23	&2125	&1.03 \\
stanfordcars \cite{KrauseStarkDengFei-Fei_3DRR2013}   &8144	&196	&41.6	&2.83 \\
\hline
\end{tabular}}
\end{center}
\caption{Target datasets.}
\label{tab:source_dataset}
\end{table}

\section{Additional Experiments}
\label{apx:additional-experiments}

\begin{figure}[b]
    \centering
    \includegraphics[width=.40\linewidth]{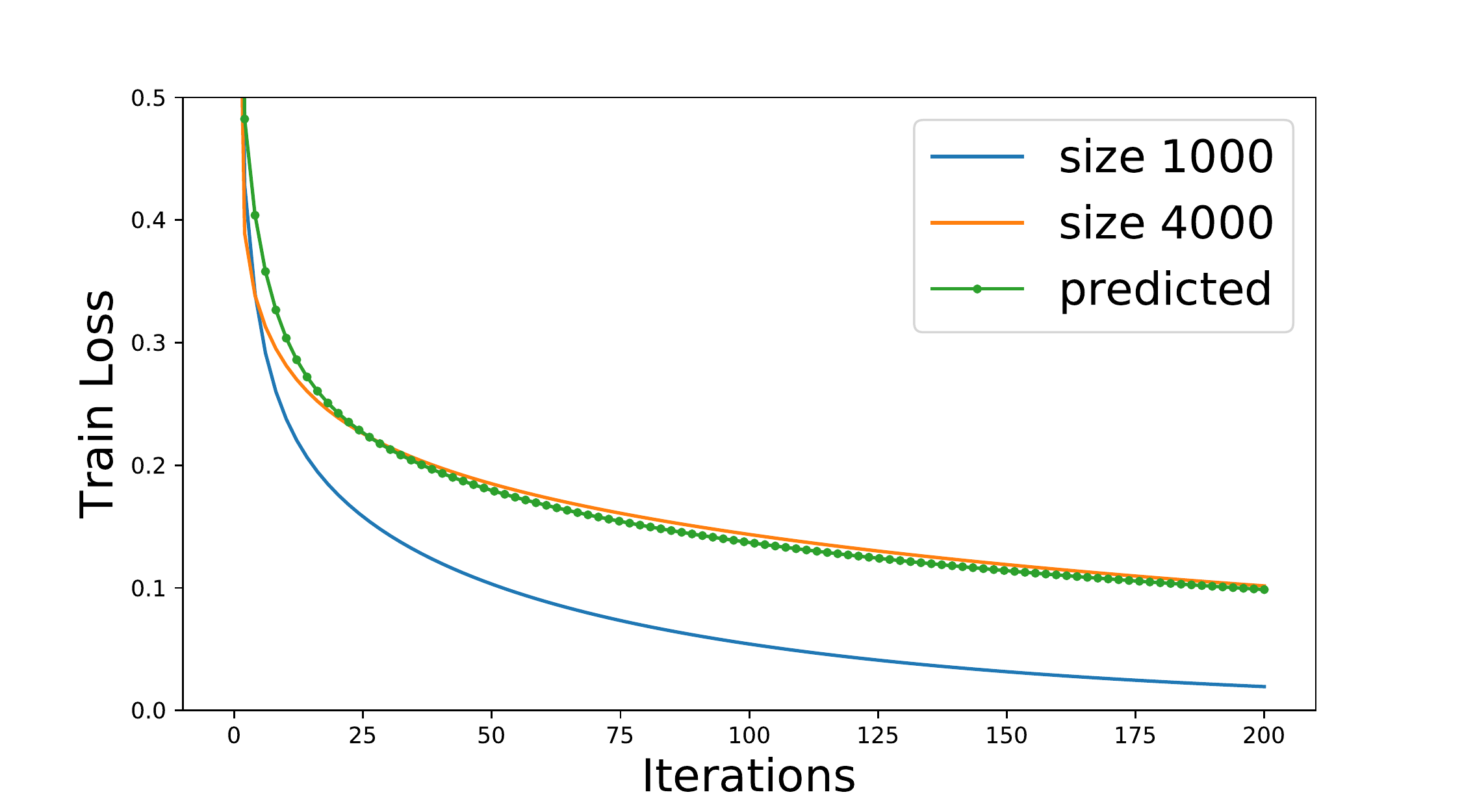}  %
    \includegraphics[width=.40\linewidth]{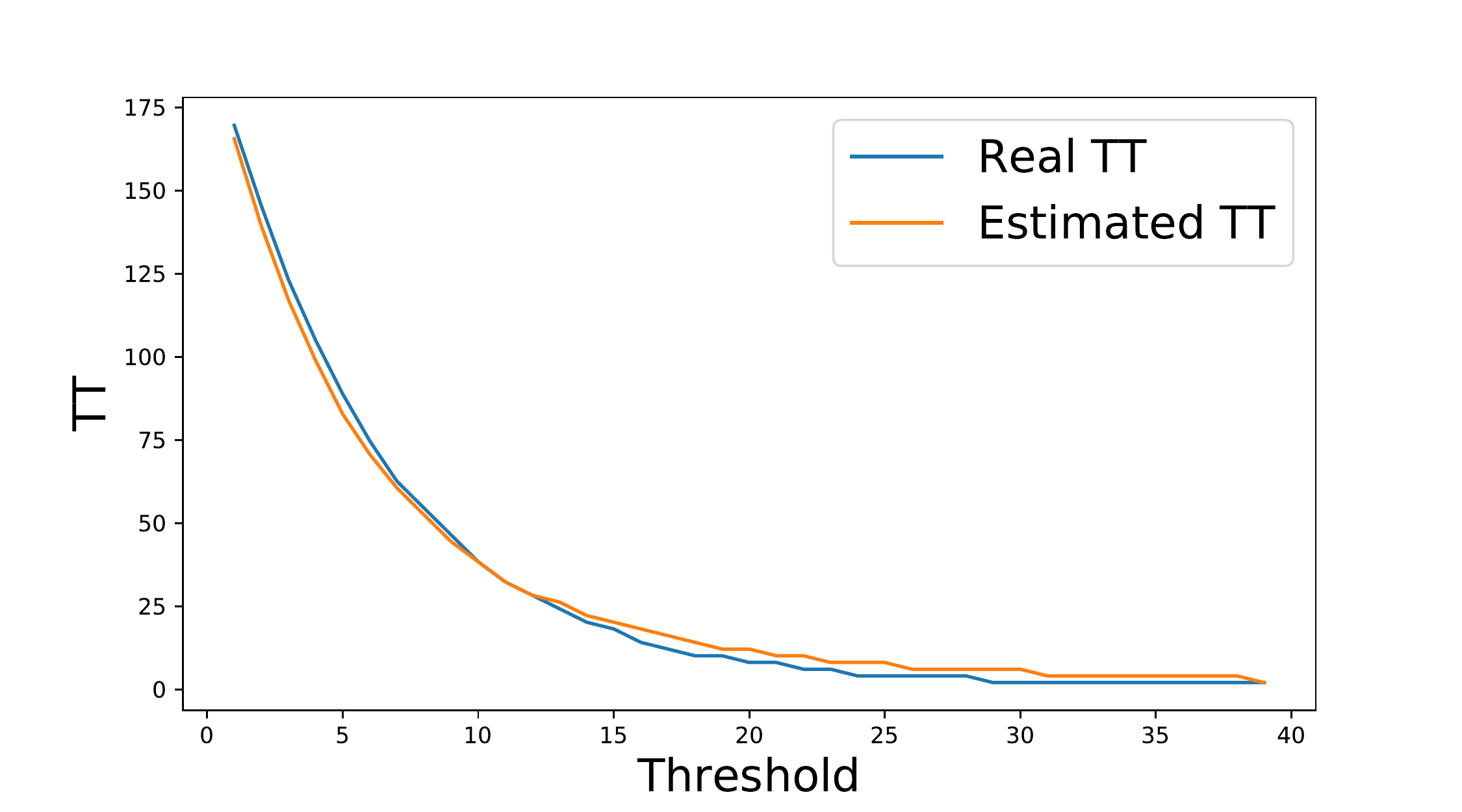}
    \caption{\textbf{Training-time prediction using a subset of the data.} \textbf{(Left)} Using the method described in \Cref{apx:larger-datasets}, we predict (green) the loss curve on a large dataset of $N=4000$ samples (orange) using a subset of $N_0=1000$ samples (blue). In \Cref{fig:dataset-size} we show a similar result using a much smaller subset of $N_0=100$ samples. \textbf{(Right)} Corresponding estimated training time on the larger dataset at different thresholds $\epsilon$ compared to the real training time on the larger dataset.}
    \label{fig:dataset-size-larger-dataset}
\end{figure}

\textbf{Prediction of training time using a subset of samples.} In \Cref{sec:computing-TT} we suggest that in the case of MSE loss, it is possible to predict the training time on a large dataset using a smaller subset of samples (we discuss the details in \Cref{apx:larger-datasets}). In \Cref{fig:dataset-size-larger-dataset} we show the result of predicting the loss curve on a dataset of $N=4000$ samples using a subset of $N=1000$ samples. Similarly, in \Cref{fig:dataset-size} (top row) we show the more difficult example of predicting the loss curve on $N=1000$ samples using a very small subset of $N_0=100$ samples. In both cases we correctly predict that training on a larger dataset is slower, in particular we correctly predict the asymptotic convergence phase. Note in the case $N_0=100$ the prediction is less accurate, this is in part due to the eigenspectrum of $\Theta$ being still far from its limiting behaviour achieved for large number of data (see \Cref{apx:larger-datasets}).

\begin{figure}
    \centering
    \includegraphics[width=.28\linewidth]{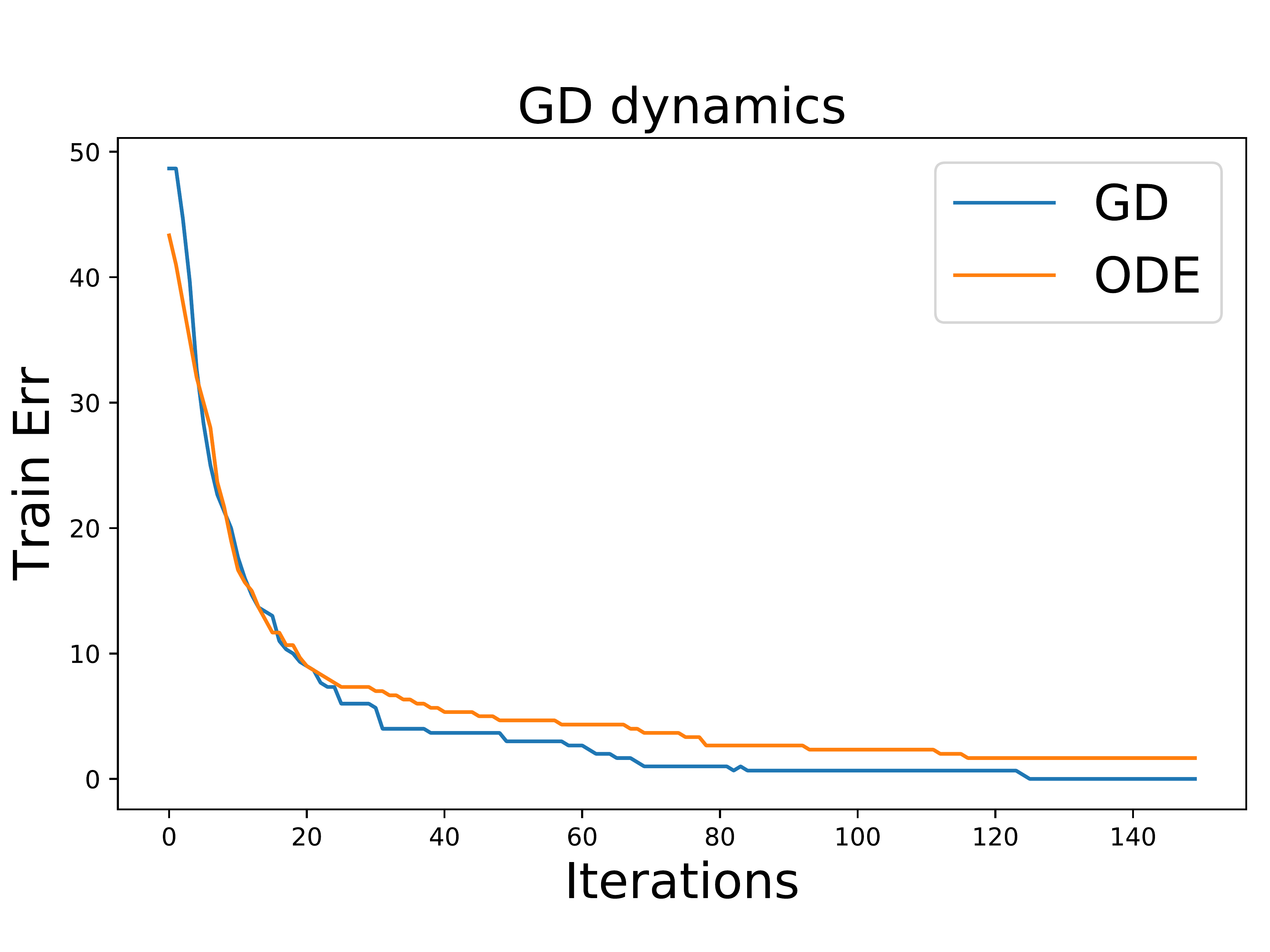}
    \includegraphics[width=.28\linewidth]{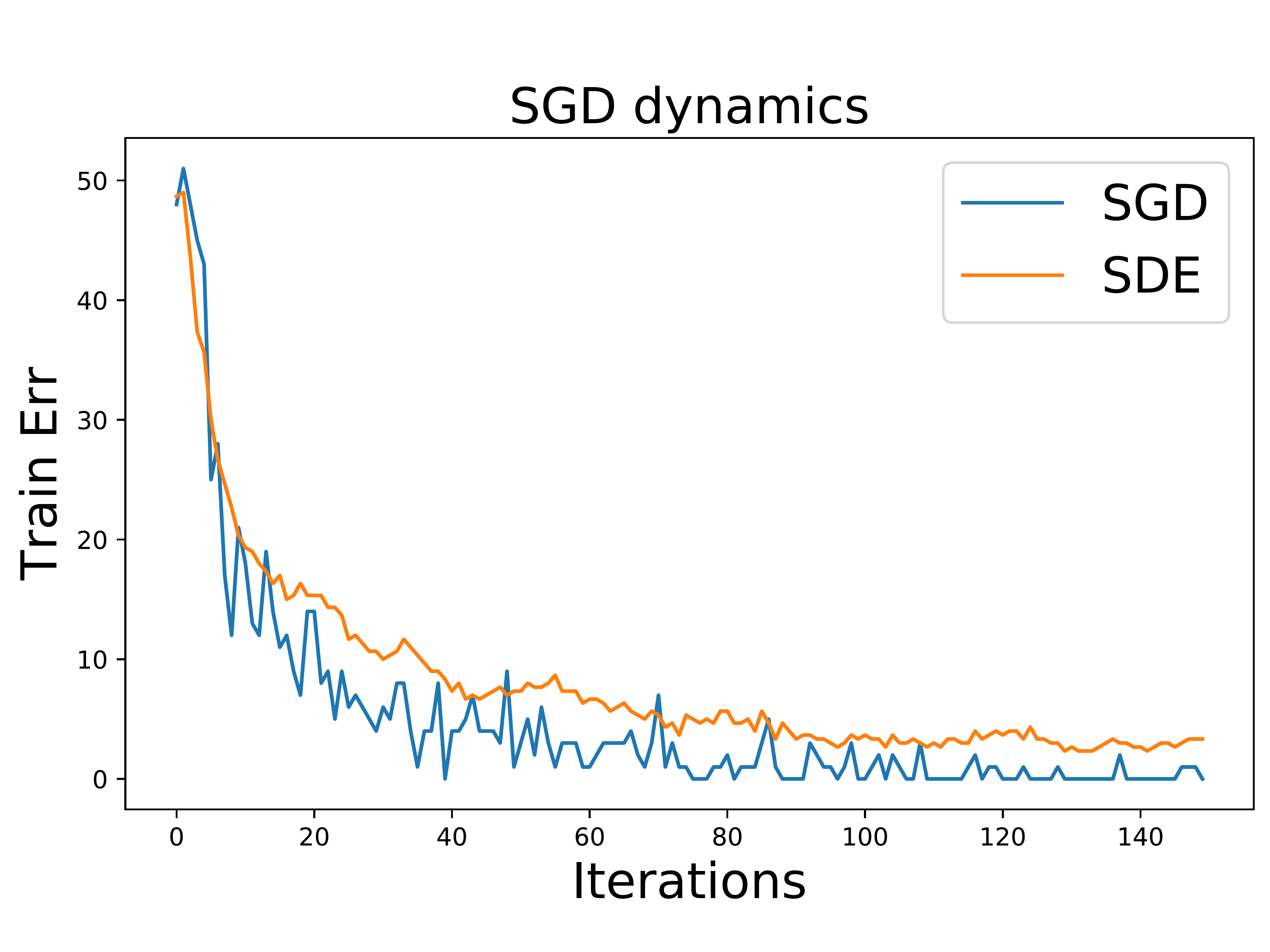}
    \includegraphics[width=.29\linewidth]{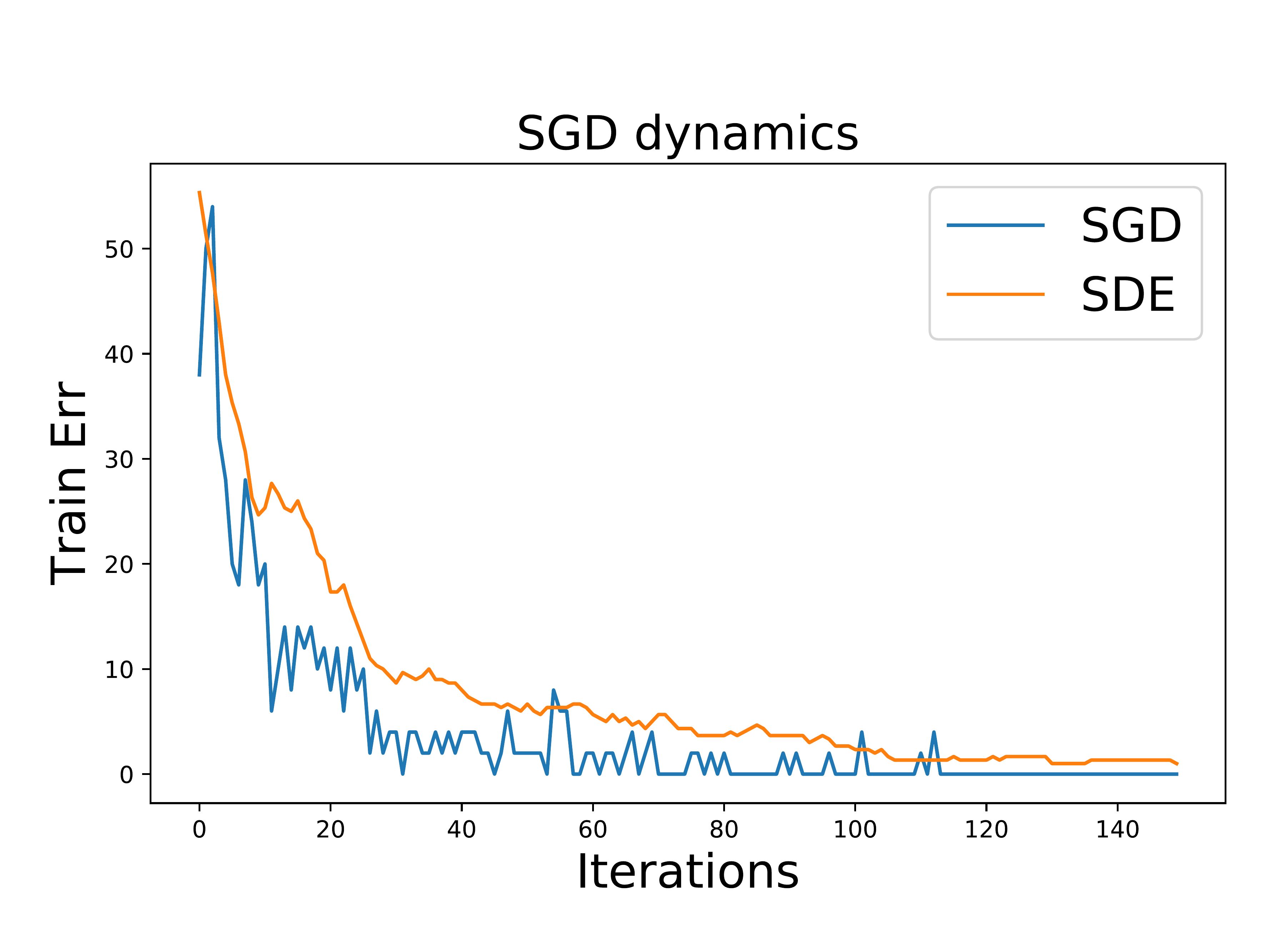}
    \caption{\textbf{(Left)} Comparison of the real error curve on CIFAR10 using gradient descent and the predicted curve. \textbf{(Center)} Same as before, but this time we train using SGD and compare it with the prediction using the technique described in \Cref{sec:computing-TT} to approximate the covariance of the SGD noise that appears in the SDE in \cref{eq:activations-sde}. \textbf{(Right)} Same as (center), but using constant noise instead of rescaling the noise using the value of the loss function as described in \Cref{sec:computing-TT}. Note that in this case we do not capture the right asymptotic behavior of SGD.}
    \label{fig:GD-vs-SGD-ODE-vs-SDE}
\end{figure}

\textbf{Comparison of predicted and real error curve.} In \Cref{fig:GD-vs-SGD-ODE-vs-SDE} we compare the error curve predicted by our method and the actual train error of the model as a function of the number of optimization steps. The model is trained on a subset of 2 classes of CIFAR-10 with 150 samples. We run the comparison for both gradient descent (left) and SGD (right), using learning rate $\eta=0.001$, momentum $m=0$ and (in the case of SGD) batch size 100. In both cases we observe that the predicted curve is reasonably close to the actual curve, more so at the beginning of the training (which is expected, since the linear approximation is more likely to hold). We also perform an ablation study to see the effect of different approximation of SGD noise in the SDE in \cref{eq:activations-sde}. In  \Cref{fig:GD-vs-SGD-ODE-vs-SDE} (center) we estimate the variance of the noise of SGD at the beginning of the training, and then assume it is constant to solve the SDE. Notice that this predicts the wrong asymptotic behavior,  in particular the predicted error does not converge to zero as SGD does. In \Cref{fig:GD-vs-SGD-ODE-vs-SDE} (right) we rescale the noise as we suggest in \Cref{sec:computing-TT}: once the noise is rescaled the SDE is able to predict the right asymptotic behavior of SGD.

\begin{figure}[b]
    \centering
    \includegraphics[width=.4\linewidth]{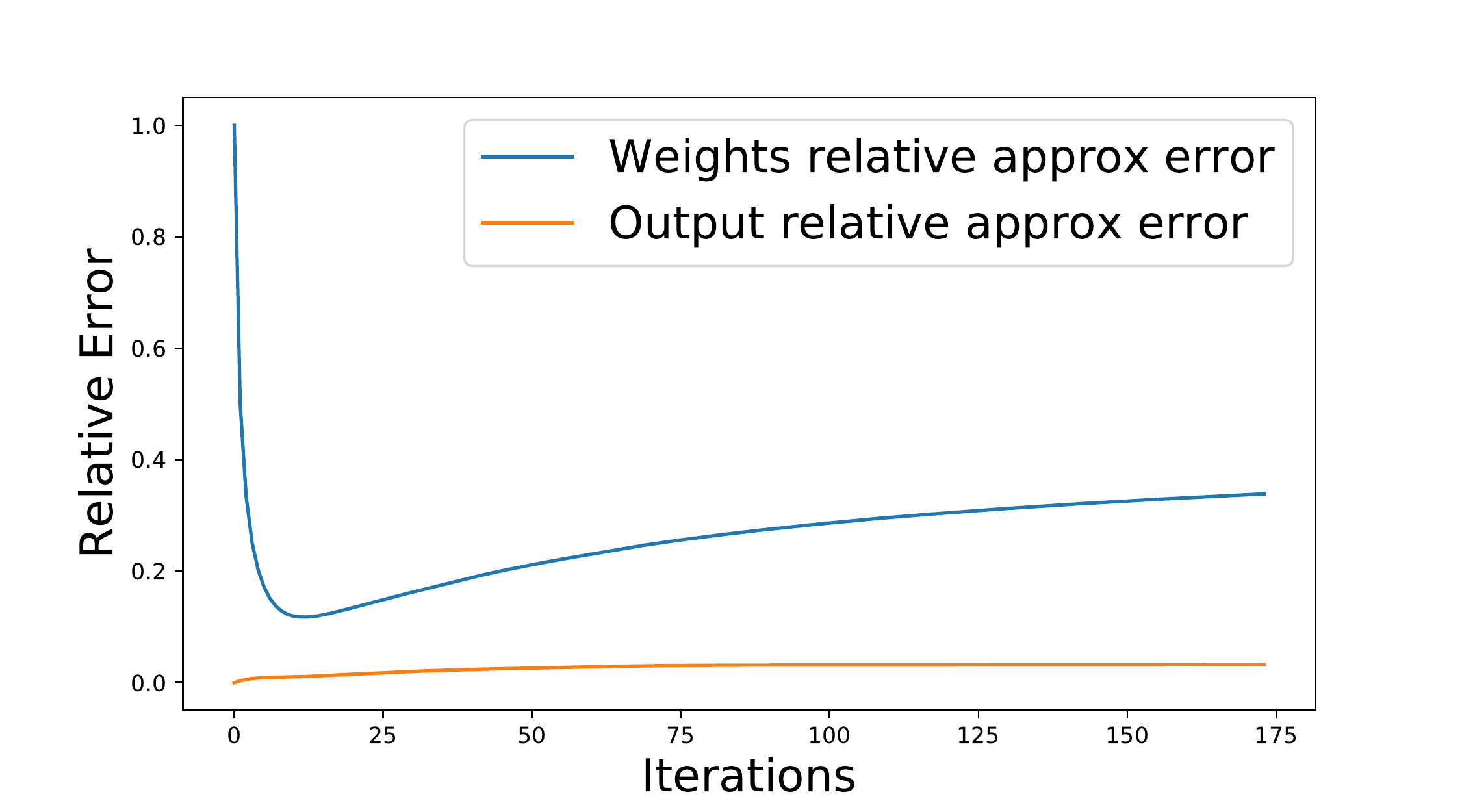}
    \caption{\textbf{Comparison of prediction accuracy in weight space vs. function space.} We compare the result of using the deterministic part of \cref{eq:activations-sde} to predict the weights $w_t$ at time $t$ and the outputs $f_t(\X)$ of the networks under GD. The relative error in predicting the outputs is much smaller than the relative error of predicting the weights at all times. This, together with the computational advantage, motivates the decision of using \cref{eq:activations-sde} to predict the behavior in function space.
    }
    \label{fig:GD-vs-SGD-alldatasets}
\end{figure}

\textbf{Prediction accuracy in weight space and function space.} In \Cref{sec:sde} and \Cref{sec:assumptions} we argue that using a differential equation to predict the dynamics in function space rather than weight space is not only faster (in the over-parametrized case), but also more accurate. In \Cref{fig:GD-vs-SGD-alldatasets} we show empirically that solving the corresponding ODE in weight space leads to a substantially larger  prediction error.

\begin{figure}
\centering
    \includegraphics[width=.99\linewidth]{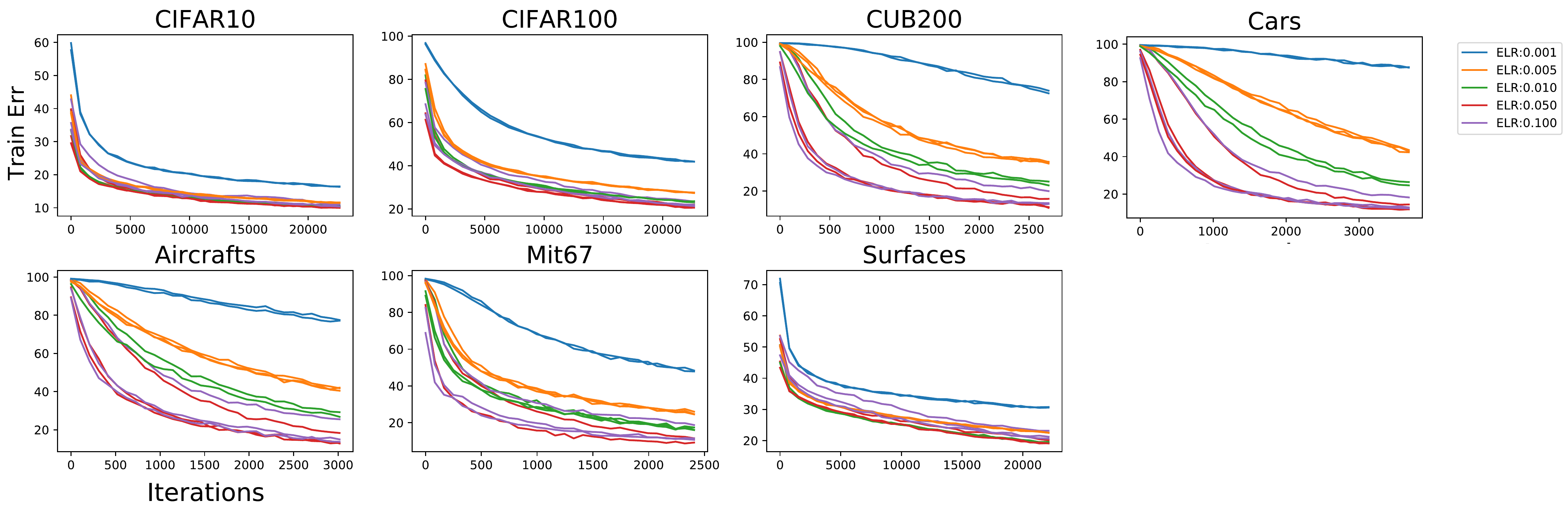}
    \caption{\textbf{Additional experiments on the effective learning rate.} We show additional plots showing the error curves obtained on different datasets using different values of the effective learning rate $\tilde{\eta}=\eta/(1-m)$, where $\eta$ is the learning rate and $m$ is the momentum. Each line is the observed error curve of a model trained with a different learning rate $\eta$ and momentum $m$. Lines with the same color have the same ELR $\tilde{\eta}$, but each has a different $\eta$ and $m$. As we note in \Cref{sec:effect-of-hyperparameters}, as long as $\tilde{\eta}$ remains the same, training dynamics with different hyper-parameters will have similar error curves.}
    \label{fig:ELR-alldatasets}
\end{figure}

\textbf{Effective learning rate.} In \Cref{sec:effect-of-hyperparameters} we note that as long as the effective learning rate $\tilde{\eta} = \eta/(1-m)$ remains constant, runs with different learning rate $\eta$ and momentum $m$ will have similar learning curve. We show a formal derivation in \Cref{sec:elr-proof}. In \Cref{fig:ELR-alldatasets} we show additional experiments, similar to \Cref{fig:effective-learning-rates},  on several other datasets to further confirm this point.

\begin{figure}
    \includegraphics[height=4cm]{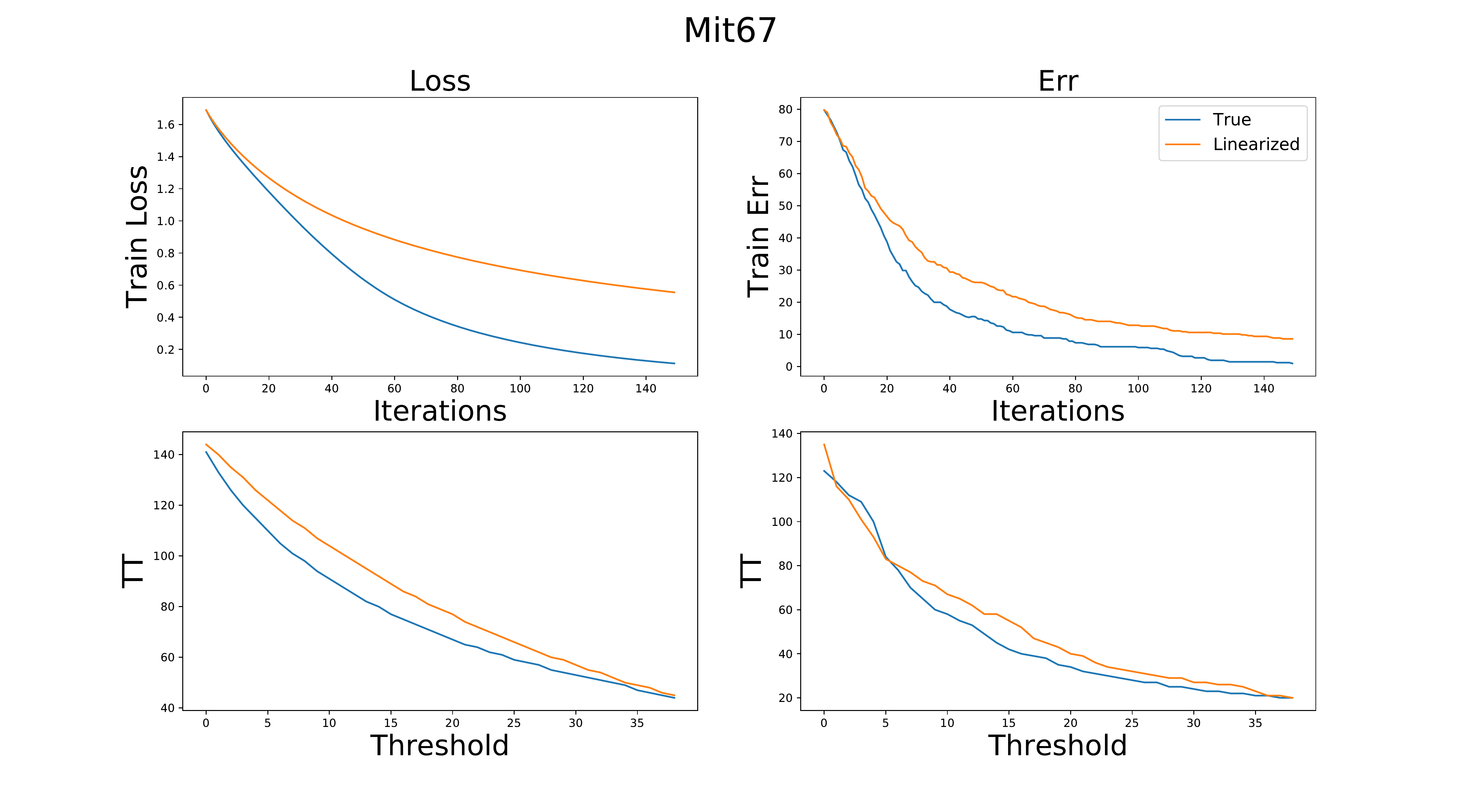}
    \includegraphics[height=4cm]{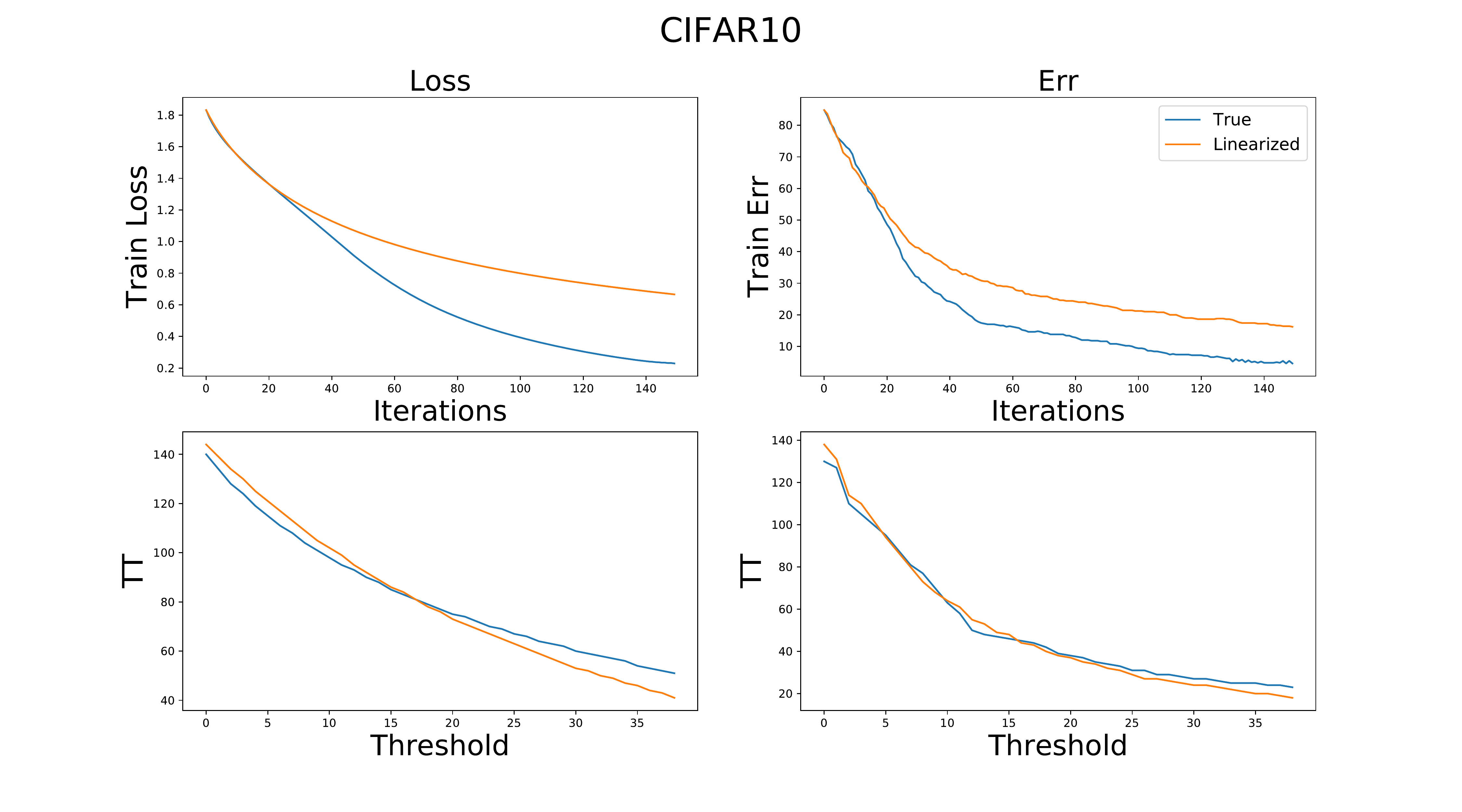}
    \caption{\textbf{Training time prediction is accurate even if loss curve prediction is not.} \textbf{(Top row)} Loss curve and error curve prediction on MIT-67 (left) and CIFAR-10 (right). \textbf{(Bottom row)} Predicted time to reach a given threshold (orange) vs real training time (blue). We note that on some datasets our loss curve prediction differs from the real curve near convergence. However, since our training time definition measures the time to reach the asymptotic value (which is what is most useful in practice) rather than the time reach an absolute threshold, this does not affect the accuracy of the prediction (see \Cref{apx:additional-experiments}).%
    }
    \label{fig:Comparing-Loss-Err-TT-predictions}
\end{figure}

\textbf{Point-wise similarity of predicted and observed loss curve.}  In some cases, we observe that the predicted and observed loss curves can differ. This is especially the case when using cross-entropy loss (\Cref{fig:Comparing-Loss-Err-TT-predictions}). We hypothesize that this may be due to improper prediction of the dynamics when the softmax output saturates, as the dynamic becomes less linear \cite{lee2019wide}. However, the train error curve (which only depends on the relative order of the outputs) remains relatively correct. We should also notice that prediction of the  $\epsilon$-training-time $\hat{T}_\epsilon$ can be accurate even if the curves are not point-wise close. The $\epsilon$-training-time seeks to find the first time after which the loss or the error is within an $\epsilon$ threshold. Hence, as long as the real and predicted loss curves have a similar asymptotic slope the prediction will be correct, as we indeed verify in  \Cref{fig:Comparing-Loss-Err-TT-predictions} (bottom).

\section{Prediction of training time on larger datasets}
\label{apx:larger-datasets}

\begin{figure}
    \centering
    \includegraphics[width=.33\linewidth]{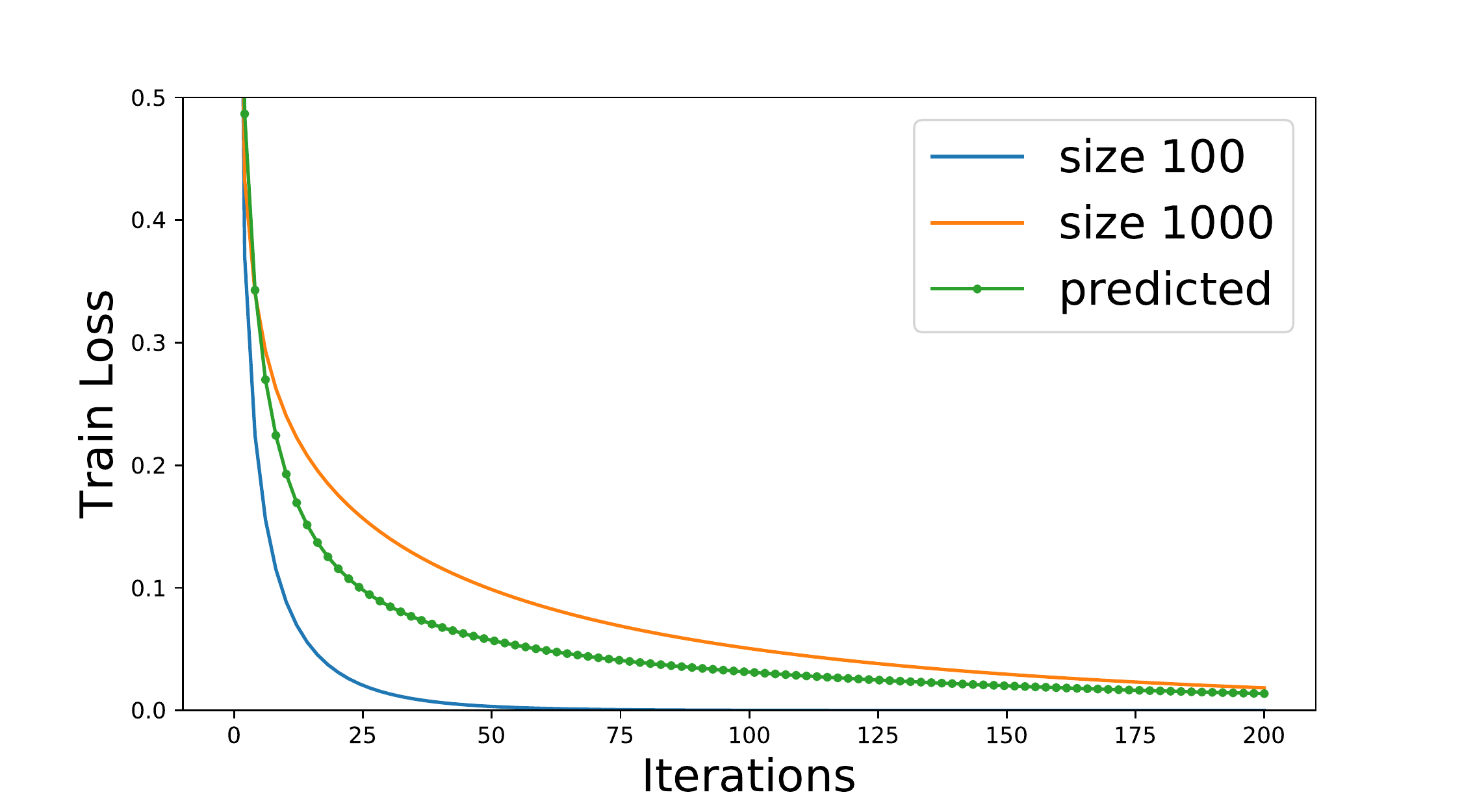}%
    \includegraphics[width=.33\linewidth]{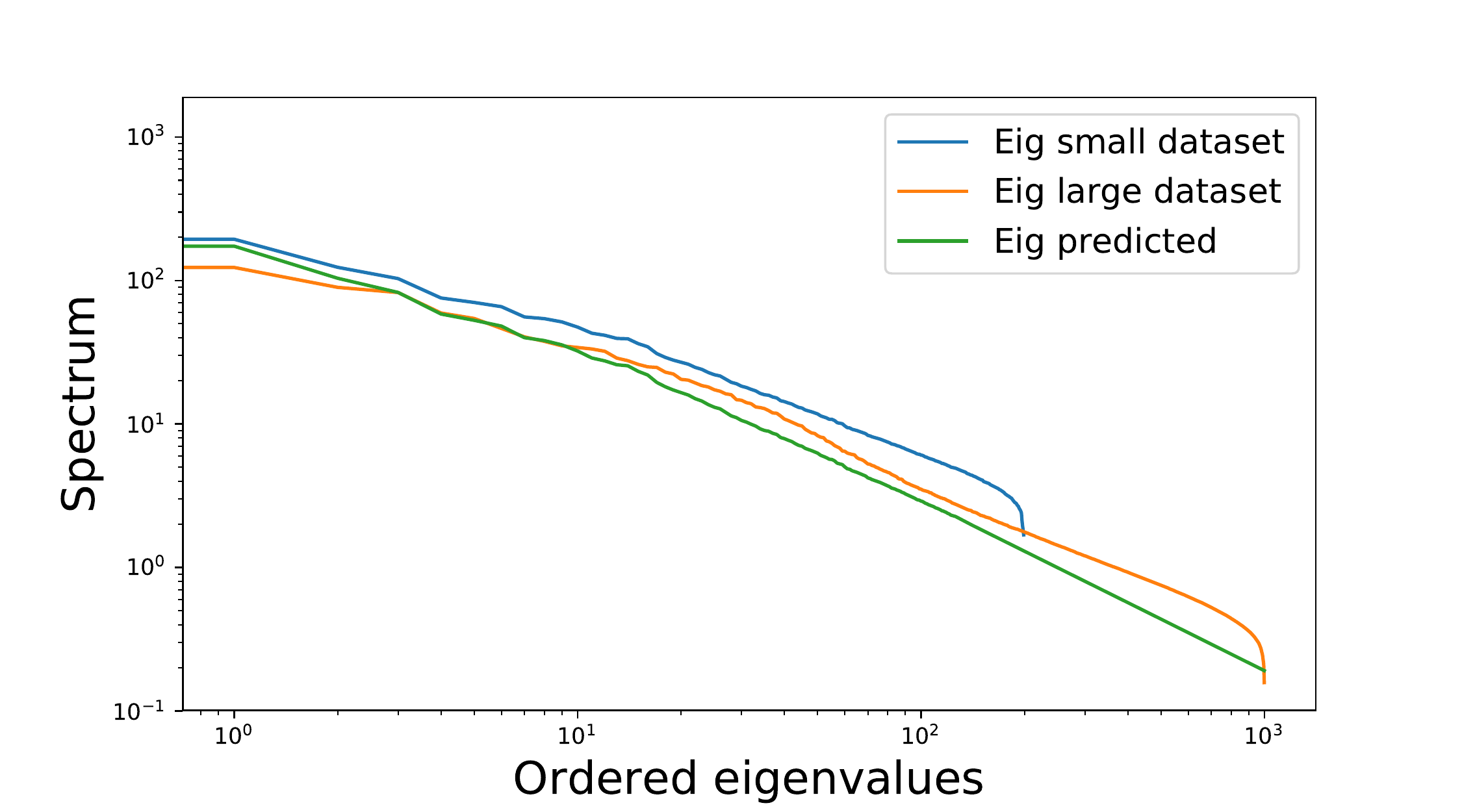}%
    \includegraphics[width=.33\linewidth]{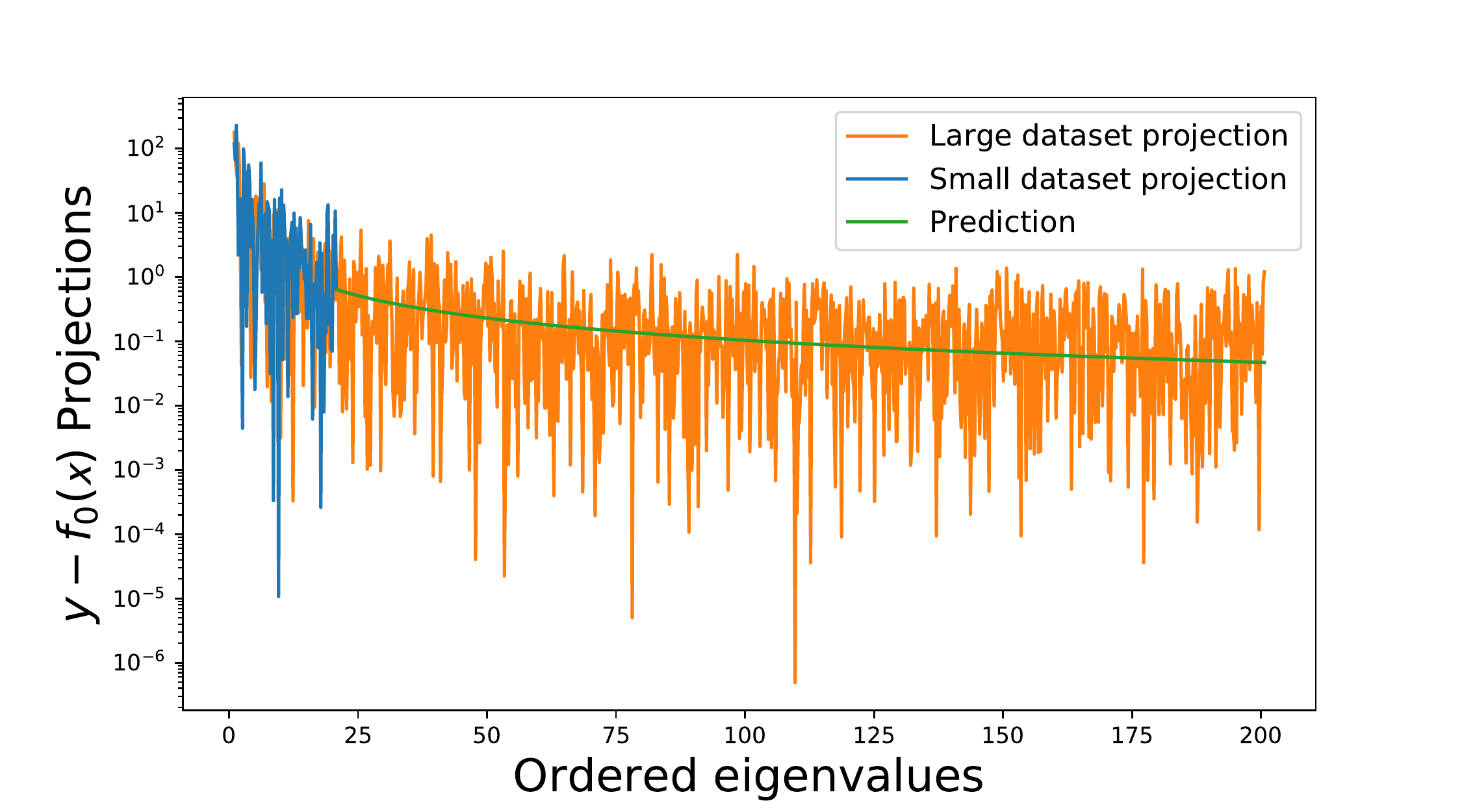}%
    \caption{\textbf{Training time prediction using a subset of the data.} \textbf{(Left)} We predict the loss curve on a large dataset of $N=1000$ samples using a subset of $N_0=100$ samples on CIFAR10 (similar results hold for other datasets presented so far). \textbf{(Center)} Eigenspectrum of $\NTK$ computed using $N_0=100$ samples (orange), $N=1000$ samples (green) and predicted spectrum using our method (blue). \textbf{(Right)} Value of the projections $p_k$ of $\dy$ on the eigenvectors of $\Theta$, computed at $N_0=100$ (orange) and $N=1000$ (blue). Note that while they approximatively follow a power-law on average, it is much more `noisy' than that of the eigenvalues. In green we show the predicted trend using our method.
    }
    \label{fig:dataset-size}
\end{figure}

In \Cref{sec:computing-TT} we suggest that, in the case of MSE loss, it is possible to predict the training time on a large dataset using a subset of the samples. To do so we leverage the fact that the eigenvalues of $\NTK$ follows a power-law which is independent on the size of the dataset for large enough sizes (see \Cref{fig:NTK-approximation}, right).  More precisely, from \Cref{prop:mse-training-time}, we know that given the eigenvalues $\lambda_k$ of $\NTK$ and the projections $p_k = \dy \cdot \v_k$ it is possible to predict the loss curve using
\[
L_t = \sum_k p_k e^{-2\eta\lambda_k t}.
\]
Let $\NTK_0$ be the Gram-matrix of the gradients computed on the small subset of $N_0$ samples, and let $\NTK$ be the Gram-matrix of the whole dataset of size $N$.
Using the fact that, as we increase the number of samples, the eigenvalues (once normalized by the dataset size) converge to a fixed limit (\Cref{fig:NTK-approximation}, right), we estimate the eigenvalues $\lambda_k$ of $\NTK$ as follow: we fit the coefficients $s$ and $c$ of a power law $\lambda_k = c k^{-s}$ to the eigenvalues of $\NTK_0$, and use the same coefficients to predict the eigenvalues of $\NTK$.
However, we notice that the coefficient $s$ (slope of the power law) estimated using a small subset of the data is often smaller than the slope observed on larger datase (note in \Cref{fig:NTK-approximation} (right) that the curves for smaller datasets are more flat). We found that using the following corrected power law increases the precision of the prediction:
\[
\hat{\lambda}_k = c k^{-s + \alpha \big(\frac{N_0}{N} -1\big)}.
\]
Empirically, we determined $\alpha \in [0.1, 0.2]$ to give a good fit over different combinations of $N$ and $N_0$. In \Cref{fig:dataset-size} (center) we compare the predicted eigenspectrum of $\NTK$ with the actual eigenspectrum of $\NTK$ .

The projections $p_k$ follow a similar power-law -- albeit more noisy (see \Cref{fig:dataset-size}, right) -- so directly fitting the data may give an incorrect result. However, notice that in this case we can exploit an additional constraint, namely that $\sum_k p_k = \|\dy\|^2$ ($\|\dy\|^2$ is a known quantity: labels and initial model predictions on the large dataset).
Let $p_k=\dy \cdot \v_k$ and let $p'_k=\dy \cdot \v_k'$ where $\v_k$ and $\v_k'$ are the eigenvectors of $\NTK$ and $\NTK_0$ respectively.
Fix a small $k_0$ (in our experiments, $k_0=100$). By convergence laws \cite{shawe2005eigenspectrum}, we have that $p'_k\simeq p_k$ when $k < k_0$. The remaining tail of $p_k$ for $k>k_0$ must now follow a power-law and also be such that $\sum_k p_k = \|\dy\|^2$. This uniquely identify the coefficients of a power law. Hence, we use the following prediction rule for $p_k$:
\[
\hat{p}_k = \begin{cases}
p'_k & \text{if } k < k_0 \\
a k^{-b} & \text{if } k \geq k_0
\end{cases}
\]
where $a$ and $b$ are such that $\hat{p}_{k_0} = p'_{k_0}$ and $\sum_k \hat{p}_k =  \|\dy\|^2$.

In \Cref{fig:dataset-size} (left), we use the approximated $\hat{\lambda}_k$ and $\hat{p}_k$ to predict the loss curve on a dataset of $N=1000$ samples using a smaller subset of $N_0=100$ samples. Notice that we correctly predict that the convergence is slower on the larger dataset. Moreover, while training on the smaller dataset quickly reaches zero, we correctly estimate the much slower asymptotic phase on the larger dataset. Increasing both $N$ and $N_0$ increases the accuracy of the estimate, since the eigenspectrum of $\Theta$ is closer to convergence: In \Cref{fig:dataset-size-larger-dataset} we show the same experiment as \Cref{fig:dataset-size} with $N_0=1000$ and $N=4000$. Note the increase in accuracy on the predicted curve.

\section{Effective learning rate}
\label{sec:elr-proof}

We now show that having a momentum term has the effect of increasing the effective learning rate in the deterministic part of \cref{eq:activations-sde}. A similar treatment of the momentum term is also in \cite[Appendix D]{Smith2018ABP}. Consider the update rule of SGD with momentum:
\begin{align*}
    a_{t+1} &= m \, a_{t} + g_{t+1}, \\
    w_{t+1} &= w_{t} - \eta\, a_{t+1},
\end{align*}
If $\eta$ is small, the weights $w_t$ will change slowly and we can consider $g_t$ to be approximately constant on short time periods, that is $g_{t+1} = g$. Under these assumptions, the gradient accumulator $a_t$ satisfies the following recursive equation:
\[
a_{t+1} = m \, a_{t} + g,
\]
which is solved by (assuming $a_0=0$ as common in most implementations):
\[
a_t = (1-m^t) \frac{g}{1-m}.
\]
In particular, $a_t$ converges exponentially fast to the asymptotic value $a^* = g/(1-m)$. Replacing this asymptotic value in the weight update equation above gives:
\[
w_{t+1} =  w_{t} - \eta a^* = w_{t} - \frac{\eta}{1-m} g = w_t - \tilde{\eta}\, g,
\]
that is, once $a_t$ reaches its asymptotic value, the weights are updated with an higher effective learning rate $\tilde{\eta} = \frac{\eta}{1-m}$. Note that this approximation remains true as long as the gradient $g_t$ does not change much in the time that it takes $a_t$ to reach its asymptotic value. This happens whenever the momentum $m$ is small (since $a_t$ will converge faster), or when $\eta$ is small ($g_t$ will change more slowly). For larger momentum and learning rate, the effective learning rate may not properly capture the effect of momentum.

\section{Proof of theorems}

\subsection{Proposition 1: SDE in function space for linearized networks trained with SGD}

We now prove our \Cref{prop:activations-sde} and show how we can approximate the SGD evolution in function space rather than in parameters space.
We follow the standard method used in \cite{hayou2019mean} to derive a general SDE for a DNN, then we speciaize it to the case of linearized deep networks.
Our notation follows \cite{lee2019wide}, we define $f_{\theta_t}(\mathcal{X}) = vec([f_t(x)]_{x\in \mathcal{X}}) \in \mathbb{R}^{CN}$ the stacked vector of model output logits for all examples, where $C$ is the number of classes and $N$ the number of samples in the training set.

To describe SGD dynamics in function space we start from deriving the SDE in parameter space.
In order to derive the SDE required to model SGD we will start describing the discrete update of SGD as done in \cite{hayou2019mean}.
\begin{equation}
    \label{eq:discrete_SGD}
    \theta_{t+1} = \theta_t - \eta \nabla_\theta \mathcal{L}^B(\theta_t)
\end{equation}
where $\mathcal{L}^B(\theta_t) = \mathcal{L}(f_{\theta_t}(\mathcal{X}^B), \mathcal{Y}^B)$ is the average loss on a mini-batch $B$ (for simplicity, we assume that $B$ is a set of indexes sampled with replacement).

The mini-batch gradient $\nabla_\theta \mathcal{L}^B(\theta_t)$ is an unbiased estimator of the full gradient, in particular the following holds:
\begin{equation}
    \E[\nabla_\theta \mathcal{L}^{B} (\theta_t)] =0 \qquad \cov[\nabla_\theta \mathcal{L}^{B} (\theta_t)] = \frac{\Sigma (\theta_t)}{|B|}
\end{equation}
Where we defined the covariance of the gradients as:
\begin{equation*}
    \Sigma(\theta_t) :=  \E \big[ (g_i \nabla_{f_t(x_i)} \L) \otimes (g_i \nabla_{f_t(x_i)} \L) \big] - \E\big[g_i \nabla_{f_t(x_i)} \L\big] \otimes \E\big[g_i \nabla_{f_t(x_i)} \L\big]
\end{equation*}
and $g_i := \nabla_w f_0(x_i)$. The first term in the covariance is the second order moment matrix while the second term is the outer product of the average gradient.

Following standard approximation arguments (see \cite{DBLP:journals/corr/abs-1710-11029} and references there in) in the limit of small learning rate $\eta$ we can approximate the discrete stochastic equation \cref{eq:discrete_SGD}  with the SDE:
\begin{equation}
    d\theta_t = - \eta \nabla_\theta \mathcal{L}(\theta_t)dt + \frac{\eta}{\sqrt{|B|}} \Sigma(\theta_t)^\frac{1}{2} dn
\end{equation}
where $n(t)$ is a Brownian motion.

Given this result, we are going now to describe how to derive the SDE for the output $f_t(\X)$ of the network on the train set $\X$. Using Ito's lemma (see \cite{hayou2019mean} and references there in), given a random variable $\theta$ that evolves according to an SDE, we can obtain a corresponding SDE that describes the evolution of a function of $\theta$. Applying the lemma to $f_\theta(\X)$ we obtain:
\begin{equation}
    df_t(\mathcal{X}) = [- \eta \NTK_t \nabla_{f_t} \mathcal{L}(f_t(\mathcal{X}), \mathcal{Y}) + \frac{1}{2}vec(A)] dt +\frac{\eta}{\sqrt{|B|}} \nabla_\theta f(\mathcal{X}) \Sigma(\theta_t)^\frac{1}{2} dn
\end{equation}
where $\nabla_\theta f(\mathcal{X}) \in \mathbb{R}^{CN \times D}$ is the jacobian matrix and $D$ is the number of parameters.
Note $A$ is a $N \times C$ matrix which, denoting by $f^{(j)}_\theta(x)$ the $j$-th output of the model on a sample $x$, is given by:
\[
A_{ij} = \operatorname{tr}[\Sigma(\theta_t) \nabla^2_\theta f^{(j)}_\theta(x_{i})].
\]
Using the fact that in our case the model is linearized, so $f_\theta(x)$ is a linear function of $\theta$, we have that $\nabla^2_\theta f^{(j)}(x) = 0$ and hence $A=0$. This leaves us with the SDE:
\begin{equation}
    df_t(\mathcal{X}) = - \eta \NTK_t \nabla_{f_t} \mathcal{L} dt + \frac{\eta}{\sqrt{|B|}} \nabla_\theta f(\mathcal{X}) \Sigma(\theta_t)^\frac{1}{2} dn
\end{equation}
as we wanted.

\subsection{Proposition 2: Loss decomposition} \label{sec:loss-decomposition}
Let $\nabla_w f_w(\X) = V \Lambda U$  be the singular value decomposition of $\nabla_w f_w(\X)$ where $\Lambda$ is a rectangular matrix (of the same size of $\nabla_w f_w(\X)$) containing the singular values $\{\sigma_1, \ldots, \sigma_N\}$ on the diagonal. Both $U$ and $V$ are orthogonal matrices. Note that we have
\begin{align*}
    S &= \nabla_w f_w(\X)^T \nabla_w f_w(\X) = U^T \Lambda^T \Lambda U,\\
    \NTK &= \nabla_w f_w(\X) \nabla_w f_w(\X)^T = V \Lambda \Lambda^T V^T.
\end{align*}
We now use the singular value decomposition to derive an expression for $\mathcal{L}_t$ in case of gradient descent and MSE loss (which we call $L_t$). In this case, the differential equation \cref{eq:activations-sde} reduces to:
\[
\dot{f}_t^\lin(\X) = - \eta \Theta (\Y - f_t^\lin(\X)),
\]
which is a linear ordinary differential equation that can be solved in closed form. In particular, we have:
\[
f_t^\lin(\X) = (I - e^{-\eta \Theta t}) \Y + e^{-\eta \Theta t} f_0(\X).
\]
Replacing this in the expression for the MSE loss at time $t$ we have:
\begin{align*}
L_t &= \sum_{i} (y_i - f_t^\lin(x_i))^2\\
&= (\Y - f_t^\lin(\X))^T (\Y - f_t^\lin(\X))\\
&= (\Y - f_0(\X))^T e^{-2 \eta \Theta t} (\Y - f_0(\X)).
\end{align*}
Now recall that, by the properties of the matrix exponential, we have:
\[
e^{-2 \eta \Theta t} = e^{-2 \eta V \Lambda \Lambda^T V^T t} = V  e^{-2 \eta \Lambda \Lambda^T t} V^T,
\]
where $e^{-2 \Lambda \Lambda^T t} = \operatorname{diag}(e^{-2\eta \lambda_1 t}, e^{-2\eta \lambda_2 t}, \ldots)$ with $\lambda_k := \sigma_k^2$. Then, defining $\dy = \Y -  f_0(\X)$ and denoting with $\v_k$ the $k$-th column of $V$ we have:
\begin{align*}
L_t &= \dy^T V e^{-2 \eta \Lambda \Lambda^T t} V^T \dy \\
&= \sum_{k=1}^N e^{-2 \eta \lambda_k t} (\dy \cdot \v_k).
\end{align*}
Now let $\u_k$ denote the $k$-th column of $U^T$ and $g_i$ the $i$-th column of $\nabla_w f_w(\X)^T$ (that is, the gradient of the $i$-th sample). To conclude the proof we only need to show that $\lambda_k \v_k = (g_i \cdot \u_k)_{i=1}^N$. But this follows directly from the SVD decompostion $\nabla_w f_w(\X) = V \Lambda U$, since then $V \Lambda = \nabla_w f_w(\X) U^T$.

\end{document}